\documentclass[journal, 10pt]{IEEEtran}
\IEEEoverridecommandlockouts
\usepackage{cite}
\usepackage[utf8]{inputenc} % allow utf-8 input
\usepackage[T1]{fontenc}    % use 8-bit T1 fonts
\usepackage{hyperref}       % hyperlinks
\usepackage{url}            % simple URL typesetting
\usepackage{booktabs}       % professional-quality tables
\usepackage{amsfonts}       % blackboard math symbols
\usepackage{nicefrac}       % compact symbols for 1/2, etc.
\usepackage{microtype}      % microtypography
\usepackage{graphicx}
\usepackage{hyperref}
\usepackage{multirow}
\usepackage{amsmath}
\usepackage{enumitem}   
\usepackage{adjustbox}
\usepackage{dblfloatfix}
\usepackage{xcolor}

\usepackage[nomargin,inline,draft]{fixme}
\fxsetup{theme=color,mode=multiuser}
\FXRegisterAuthor{j}{jla}{\color{purple}JLA}
\FXRegisterAuthor{r}{rdm}{\color{teal}RdM}

\def\BibTeX{{\rm B\kern-.05em{\sc i\kern-.025em b}\kern-.08em
    T\kern-.1667em\lower.7ex\hbox{E}\kern-.125emX}}

\begin{document}

\title{On the Inclusion of Spatial Information for Spatio-Temporal Neural Networks}

\author{\IEEEauthorblockN{Rodrigo de Medrano, José L. Aznarte} \\
\IEEEauthorblockA{\textit{Artificial Intelligence Department} \\
\textit{Universidad Nacional de Educación a Distancia --- UNED}\\
Madrid, Spain \\
\texttt{rdemedrano@dia.uned.es}\\
\texttt{jlaznarte@dia.uned.es}}
}

\maketitle

\begin{abstract}
  When confronting a spatio-temporal regression, it is sensible to feed the
  model with any available \textit{prior} information about the spatial
  dimension. For example, it is common to define the architecture of neural
  networks based on spatial closeness, adjacency, or correlation. A common
  alternative, if spatial information is not available or is too costly to
  introduce it in the model, is to learn it as an extra step of the model. While
  the use of \textit{prior} spatial knowledge, given or learnt, might be
  beneficial, in this work we question this principle by comparing traditional
  forms of convolution-based neural networks for regression with their respective spatial
  agnostic versions. Our results show that the typical inclusion of
  \textit{prior} spatial information is not really needed in most cases. In
  order to validate this counterintuitive result, we perform thorough
  experiments over ten different datasets related to sustainable mobility and
  air quality, substantiating our conclusions on real world problems with direct
  implications for public health and economy.
 % When confronting a spatio-temporal regression, the spatial dimension usually
 % makes use of \textit{prior} information given to the model. Concretely, neural
 % network methods generally define their architectures based on spatial
 % closeness, adjacency, or correlation. If not, these spatial relations are
 % typically learned as an extra step of the model. While the use of
 % \textit{prior} spatial knowledge, learnt or established, might be beneficial,
 % through this work we discuss this issue by comparing a naive approach that
 % makes no use of spatial knowledge, called Spatial Agnostic Neural Networks,
 % with state of the art models. We show that the typical inclusion of
 % \textit{prior} spatial information is not really needed in most cases. In order
 % to validate this counterintuitive result, we select ten different datasets
 % related to sustainable mobility and air quality which let us base our
 % conclusions in a more general frame while focus on real data and fundamental
 % fields for public health and economy.
\end{abstract}


\section{Introduction}
\label{S1}

Convolutional neural networks (CNN) are well known for their ability to handle
spatial data in several contexts, like images or spatial phenomena. However, in
the last few years they have demonstrated to hold a good position also when
dealing with temporal data. Thus, they are widely used in spatio-temporal
regression problems, with outstanding behavior when coping with both spatial and
temporal dimensions.

Due to its parameter structure, CNNs are usually employed when it is possible to
order input data in a grid. Furthermore, they treat each location equally,
learning and sharing the same weights for all spatial points. Given that it is
not rare that the phenomenon under study presents the same nature all over the
grid, in a wide range of applications this property is a clear advantage in
order to minimize the number of parameters and calculations for learning an
specific task. This leads to good performance with fewer resources compared to
feedforward neural networks (FNN) and recurrent neural networks (RNN). For
example, pollution and traffic regression share an approximately equivalent temporal
behavior and distribution at each location (at least in a close environment),
meaning that it is possible to share parameters and get a smooth approximation
for these phenomena via traditional CNNs.

However, this property of CNNs (which is usually known as
\textit{equivariance}), might not always be the best deal when solving some
typical problems: sometimes, although similar, treating all locations equally
does not hold as a valid or acceptable hypothesis and so, learning a spatial
shared-based representation might not be the best option if the system
representation is not chosen carefully. In the previous example, it is obvious
that different traffic sensors or pollution stations will have different
properties, even though their temporal dynamic will be somehow similar. For
spatio-temporal regression specifically, several proposals have been made in
order to tackle this problem, but two of them stand out for their wide
acceptance:
\begin{itemize}
\item Order the grid by Euclidean distance (from now on, just closeness) and
  use CNNs.
\item Define the system in a graph structure and model it via graph
  convolutional networks (GCN).
\end{itemize}

In both cases, a classical assumption is made: closer locations have similar
properties and, because of that, the shared-weights learned by the networks are
more reliable. This way, the spatial dimension in CNNs keeps a low number of
parameters.

However, these solutions do have some disadvantages. First, they do not
completely solve the fact that each location, although related to the rest, has
its own properties. Even more, although the assumption that closer locations
behave similarly is usually blindly accepted, this might not hold always for
real problems: not only depends on the phenomenon, but also on the temporal and
spatial granularity with which the data is taken. Thus, the benefits of learning
a latent representation based on sharing parameters are conditioned by the
particularities of each specific problem and, contrary to popular belief, the
spatial proximity between locations is not necessarily the main factor. Second,
in both cases it is necessary to introduce \textit{prior} spatial knowledge to
the system, making it less 'intelligent' and more laborious to work with.

In this paper we focus on whether defining adjacency-based convolutional
architectures for regression problems is as important or positive as 
has traditionally been assumed. To explore and contrast our hypothesis,
we propose to compare a set of widely used traditional convolutional 
methods with their respective spatial agnostic versions. Here, the denotation
``spatial agnostic'' makes reference at not including specific mechanisms
that exploit spatial information explicitly. By showing that no improvement
is reported when using \textit{prior} spatial knowledge, we can reject the 
idea that models with a spatial bias will result systematically in 
better forecasters.
% \jnote{No entiendo esta última frase.} 
Also, models with spatial agnostic nature can be a suitable choice
when spatial information is not easily achievable or within reach.

What happens if we closely examine the temporal dimension? In multiple real
applications in which this spatial agnosticism does not exactly hold, temporal
equivalence between locations is more plausible: temporal distributions along
spatial points might better fulfill the assumption of sharing parameters
compared to the spatial dimension. This means that, while it is common to use
some sort of recurrent module to model temporal relations, convolutions can be
perfectly valid candidates for this work, using a lower number of parameters.
Thus, sharing parameters for all locations between subsequent past timesteps in
the temporal dimension might work better than in the spatial dimension.

% \rnote{Quizás deberíamos aclarar que en nuestro entorno, los videos (computer
%   vision en general) no están incluidos como series espacio-temporales. Algún
%   revisor pidió mnist y otros videos como datasets también.}\jnote{Creo que la
%   clave está en la palabra ``regression'', en la que pondría más peso.}

To validate our hypothesis, we compare several models and dig deeper in the real
importance of closeness relations through extensive experimentation. For this
purpose, the vast field of air quality and sustainable mobility has been chosen.
With a wide number of long spatio-temporal series with spatial particularities
but approximately equivalent temporal dynamics (due to their relation with human
behavior) and high non-linearity, it is a perfect field to corroborate our
hypotheses. Since it is considered of great importance for public health and
also to economy, it is potentially beneficial to have simpler and easily
deployable models in this field.

The main contributions of this study are summarized as follows:
\begin{itemize}
\item We delve into the counterintuitive idea that including spatial relations
  based on closeness are not necessarily the optimal option when working with
  neural networks for regression in spatio-temporal problems. Concretely, we
  compare several traditional methodologies with their respective spatial
  agnostic version.
\item The contribution is illustrated by tackling a variety of prediction
  problems related to air quality and sustainable mobility. All of them are
  considered of great importance and significantly complex for both spatial and
  temporal dimensions.
\item Results show that spatial agnostic methods equal state-of-the-art models
  in accuracy without the need of \textit{prior} spatial information.
\end{itemize}

The rest of the paper is organized as follows: related work is discussed in
Section \ref{S2}, while Section \ref{S3} presents the methods and all needed
theory for this work. Then, in Section \ref{S4} we introduce our datasets,
experimental design, and its properties. Section \ref{S5} illustrates the
evaluation of the proposed architecture as derived after appropriate
experimentation. Finally, in Section \ref{S6} we point out conclusions from our
work.

% %%%%%%%%%%%% %
% RELATED WORK %
% %%%%%%%%%%%% %
\section{Related work}
\label{S2}

\subsection{The rise of convolutions}
\label{S2.1}

Since CNNs were proposed as neural architectures
\cite{lecun1989backpropagation}, they have shown to handle especially well
spatially-ordered data. During the last decades, this kind of neural networks
have grown in importance, becoming one of the most used neural paradigms for a
wide number of applications.

In the case of intrinsic 2D problems, like images, CNNs have turned out to be
the option per excellence. Concretely, with \cite{krizhevsky_imagenet_2012}
started a reign of CNN for computer vision problems. Not much later, the idea
that weight sharing could lead to potentially suboptimal performance for some
images, like portraits, was studied \cite{taigman_deepface_2014}. In the
present, CNNs are widely used for this kind of problem and have been well
characterized.

However, CNNs are not constrained to natural 2D systems. For example, time
series seen as a 1D sequence have been handled by convolutional models with good
results \cite{zhao_convolutional_2017, cui_multi-scale_2016}. Spatio-temporal
series have growth in importance and CNNs have been well studied and are already a 
standard when dealing with this kind of series 
\cite{rodrigues_beyond_2020,tu_mapping_2017}.
A similar field to spatio-temporal series is video-sequence analysis,
where both spatial and temporal relations need to be modeled \cite{nam_learning_2016}. 
Within this last topic, some examples in which parameter sharing is indeed highly 
positive can be found, as for example enhancing video spatial resolution for creating 
smooth results \cite{kappeler_video_2016}.

\subsection{Spatial dimension in spatio-temporal neural networks}
\label{S2.2}

In spatio-temporal regression specifically, convolution based networks are one
of the leading options too. As explained in Section \ref{S1}, convolution shines
in a wide range of applications involving physical spatial locations. However,
how this dimension is treated by the convolution has not received particular
attention. Thus, we have several options that are widely used but not
necessarily optimal.

For example, in traffic forecasting, defining your space as a natural grid.
\cite{jo_image--image_2019} is a good example of 2D image-to-image prediction
problem in which, by using channels as timesteps and 3D kernels, spatio-temporal
relations are exploited. As average traffic speeds for each road segment is
used, no need to \textit{prior} spatial information is needed and the grid
arrangement is natural. However, closer areas are not necessarily more related.
In \cite{guo_deep_2019} it is shown that the 3D convolution might work better,
but the same spatial arrangements and assumptions are made.

When measurement points are directly used as an arrangement for the spatial
dimension, not only it is necessary to impose same closeness supposition than
before, but a special treatment is usually needed to arrange locations
correctly. Some examples are \cite{wu_hybrid_2018}, where the authors order
traffic sensors in a 1D grid; or \cite{de_medrano_spatio-temporal_2020}, where 
measurement points are ordered as 2D images.

In recent years, graphs-based networks have received increasing attention. GCNs
not only have shown a very competitive performance, but a graph structure is
more suitable than grids for some specific problems where relations might be
non-Euclidean and directional \cite{li_diffusion_2018}. Among the different
convolutions in graphs, all of them depend heavily on an adjacency matrix which
usually needs to be manually defined. This adjacency matrix is of great
importance as it defines the graph relations and structure. Depending on the
proposal, this matrix might be defined differently: it usually is defined by
spatial closeness \cite{guo_attention_2019,zhang_graph_2020}, but there are no
restrictions. While this freedom to define the adjacency matrix might help to
avoid the closeness assumption, it would force you to find which \textit{prior}
information may be more optimal for your particular problem. If compared to
traditional CNN, GCN presents another advantage: they can naturally process
information from a $K-hop$ neighborhood \cite{zhou_graph_2017}, not restricting
themselves to uniquely adjacent nodes.

Temporal relations with neural networks are usually constrained to using some
kind of RNNs. Although much proposals have been done, through this work we will
not focus on this broad topic and we will limit its use to standards.

\subsection{Non-locally dependent proposals}
\label{S2.3}

The idea that a fixed arrangement for learning spatial relations might not be
the best deal is not new in spatio-temporal series forecasting. Lu et al
\cite{lu_modeling_2018} state that \textit{"the existence of spatial
  heterogeneity imposes great influence on modeling the extent and degree of
  road traffic correlation, which is usually neglected by the traditional
  distance based method"}, and proposed a data-driven approach to measure these
correlations. From this starting point, we can select several works that have
contributed to refine and depend less on \textit{prior} information in the
spatial dimension using neural models.

By using a hierarchical clustering over the spatio-temporal data,
\cite{asadi_spatio-temporal_2020} refines spatial relations. However, it uses a
distance matrix in the process, introducing the aforementioned bias by
closeness. In \cite{aram_spatiotemporal_2015}, a lasso methodology is used to
obtain a sparse model of the system dynamics, which simultaneously identifies
spatial correlation along with model parameters.

Attention mechanisms, which appeared on the deep learning scene a few years ago,
are a natural way to learn relations beyond the network original assumptions. In
this context, several works have used attention weights to improve performance
and demonstrate the correctness of their work with both, grid structure
\cite{do_effective_2019} and graph structure \cite{yu_forecasting_2020}.
However, \cite{de_medrano_spatio-temporal_2020} shows how closer locations are not 
necessarily more related,
and depending on the problem and the characteristics of the regression, other
considerations might be more important when learning spatial relations.

Closer to our work, in \cite{wu_connecting_2020} a similar issue but with
general multivariate time series forecasting is put on the table: existing
methods usually fail to fully exploit latent spatial dependencies between pairs
of variables and GCNs require well-defined graph structures which means they
cannot be applied directly for multivariate time series where the dependencies
are not known in advance. In their proposal, they construct a new model that
tackles both problems. \cite{uselis_localized_2020} focus its efforts on dealing
with the fact that different spatial locations might have at some degree
different dynamics by using traditional CNNs but with the introduction of
learnable local inputs/latent variables and learnable local transformations of
the inputs.

In the end, all these works focus its attention on solving a specific regression
problem, but not delve into how the spatial dimension should be really treated.
Furthermore, all these methodologies have in common the need to make their
models considerably more complex in order to overcome spatial agnosticism,
generally starting from usual convolution operators and refining themselves via
extra mechanisms or modules.

% %%%%%% %
% MODELS %
% %%%%%% %
\section{Methods}
\label{S3}

Through this section, we present all the theoretical methods and foundations in
which our study bases its ideas and experiments on the role of spatial agnosticism in
spatio-temporal series. The code for this paper is available in
\textit{\url{https://github.com/rdemedrano/SANN}}.

\subsection{Preliminaries}
\label{S3.1}

As we intend to demonstrate how the typical intrinsic spatial information given to different forms of convolutional methods is not as important as always assumed, 
we focus this paper in comparing traditional models with their respective agnostic version. Before explaining this methodologies, we introduce some general aspects.

Given a spatio-temporal sequence $X$, let us call $N$ to the total number of 
timesteps and $S$ the total number of spatial points. With this notation, a 
spatio-temporal sample from the series writes as ${x_{t_i,s_j} : i = 1, . . . , T; 
j = 1, . . . , S}$, being $T$ the total number of timesteps conforming the sample. 
$X_{t}$ is the slice of series $X$ for timestep $t$ at all locations, and $X_{T,j}$ is the 
slice of series $X$ in location $j$ for all timesteps. The predicted
series is represented by $\tilde{x}_{t'_i,s_j}: i = 1, . . . , T'; j = 1, . . . , 
S$, where $T'$ is the total number of predicted timesteps. We assume that the 
number of spatial locations is always the same for both the input and output 
series.

\begin{figure}[tbp]
\centering
\includegraphics[width=0.5\textwidth]{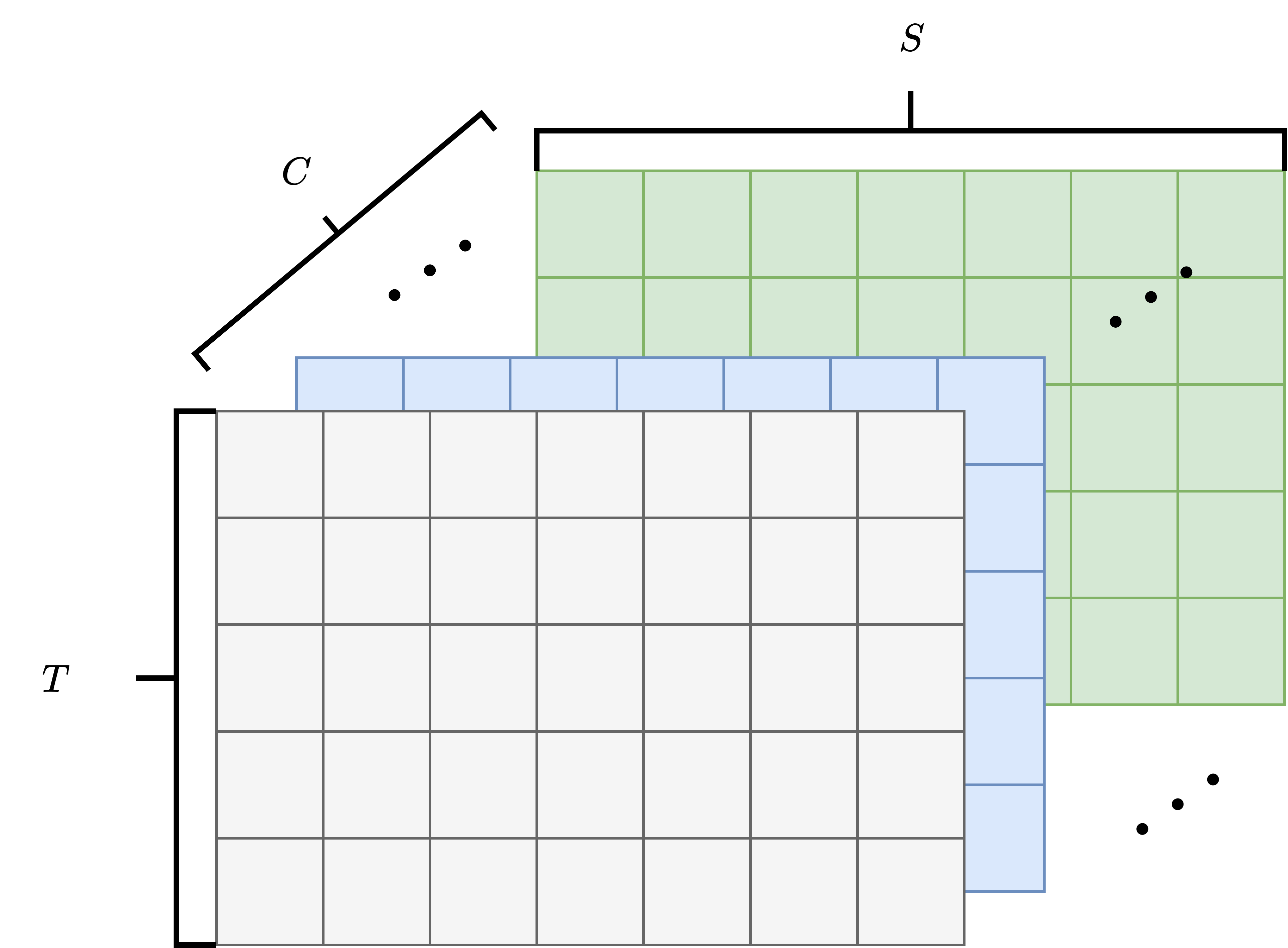}
\caption{Input sequence schematic. As long as all variables are spatio-temporal and have an 
equivalent structure for both dimensions, these sequences can be easily introduced as $C 
\times T \times S$ images, with variable, temporal and spatial dimension respectively.}
\label{fig:sann_1}
\end{figure}

For all models, the input sequence scheme relies upon a $C \times T \times S$ images 
as shown in Figure \ref{fig:sann_1}, where the number of channels $C$ represents the 
number of input spatio-temporal variables. During this paper, we will work with $C=1$ 
(the studied series by itself), but is easily extensible to any value. Also, all models 
will consist in one convolution layer outputting a $T \times S \times H$
tensor, where $H$ is the dimension of the new hidden state or number of new channels.
This approach allows us to standardize the input and output format of the convolution
layers for all methods.

\subsection{Traditional Convolutional Networks}
\label{S3.2}

Now we present the traditional format of convolution-based networks
for spatio-temporal series, and we detail how they will be used for testing our main hypothesis.

\subsubsection{Convolutional Neural Networks (CNN)}
\label{S3.2.1}

Convolutional Neural Networks are based on the idea of the convolution operation. 
Convolution itself ($*$ operator) has the following form for 2D images:

\begin{equation}
    \label{eq:1}
    (x * K)(i,j) = \sum_{m}^{k_{1}}\sum_{n}^{k_{2}} x(m,n)K(i-m, j-n) 
\end{equation}

where $K$ is the kernel. Thus, CNNs are characterized by learning a series of filters which
values depend on how adjacent elements are related.

\begin{figure}[tbp]
\centering
\includegraphics[width=0.5\textwidth]{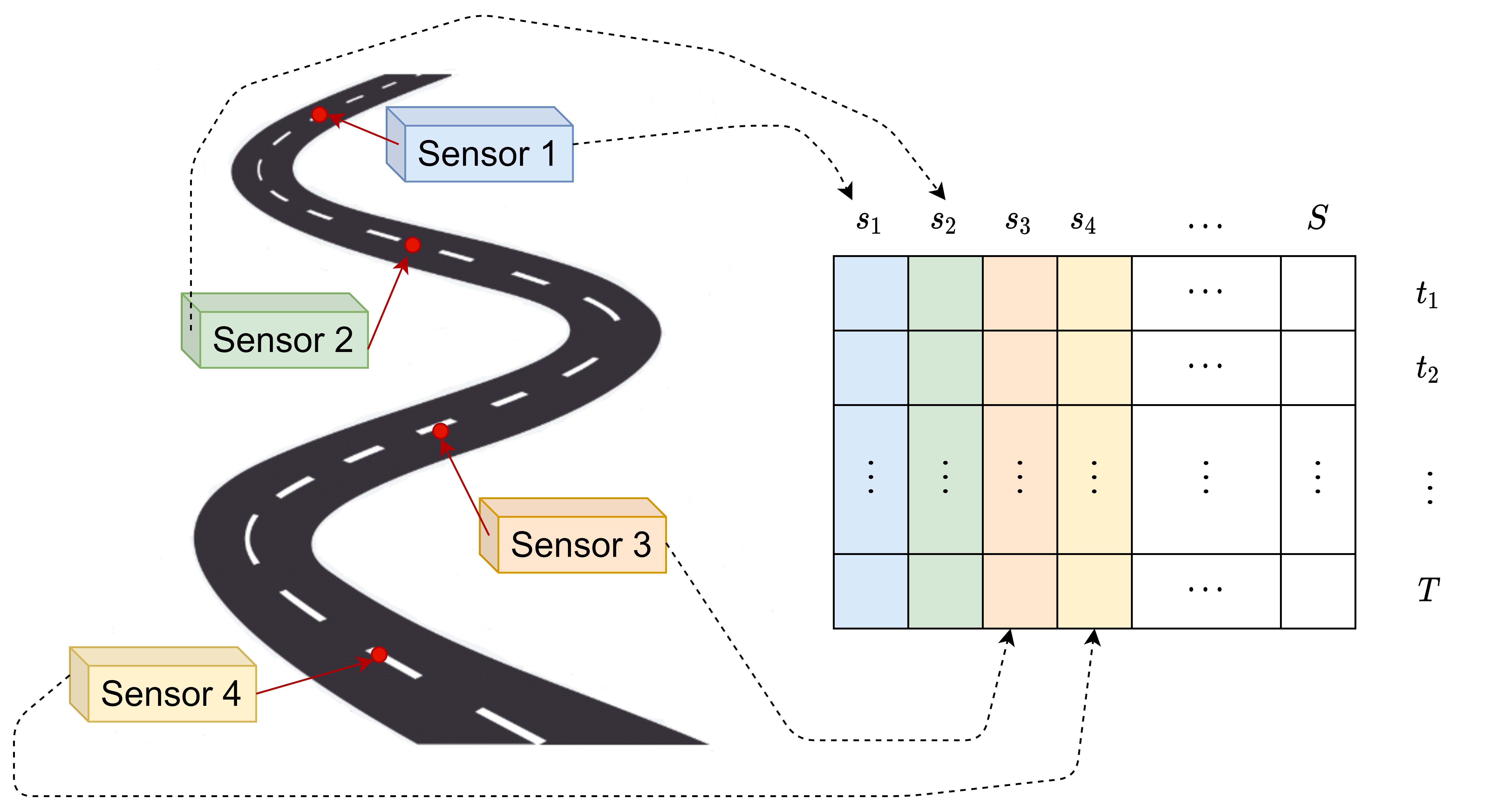}
\caption{Example of input tensor definition. Given a network of traffic sensors and
its historical series, the objective is predicting future timesteps for all locations
using CNNs. The input sequence order in the spatial dimension is usually 
defined by a logical arrangement of the relative position of the sensors in the network.
Traditionally, it is expected to improve network learning through this strategy, generating softer filters by exploiting adjacency relationships.}
\label{fig:cnn_1}
\end{figure}

In its classical form, CNNs for spatio-temporal regression rely on ordering the input
sequence by spatial adjacency or closeness. Thus, for each row of Fig. \ref{fig:sann_1}, spatial
zones are mapped into the input tensor in such a way that closer locations are closer in the
sequence. By doing so, we make sure that the learnable kernels can take advantage of this
\textit{prior} spatial information. The strategies that can be used to adapt convolutional 
networks to spatio-temporal problems by exploiting this spatial bias based on proximity are 
multiple (see Section \ref{S2.2}). In our case, the input to the network will be defined as shown
in Fig. \ref{fig:cnn_1}. Similarly, in the temporal dimension (columns) the kernel gathers adjacent
timesteps. 

\subsubsection{ConvLSTM Neural Networks}
\label{S3.2.2}

Long Short-Term Memory (LSTM) is a type of recurrent neural
network architecture usually used when
handling time series data with temporal auto-correlations. A LSTM
Neural Network consists of an input gate, an output gate, a memory cell
and a forget gate. During the training phase,
a weighted function is learned in each of the gates in order to control
how much the network “memorize” and “forget”.

Based on this model, the ConvLSTM model is a variation of 
LSTM capable of handling spatio-temporal processes. Comparing with
the original LSTM model, the input-to-state and state-to-state transitions
of the ConvLSTM cell involves convolutional operations, making it a well fit
for spatial relations. This model is governed by the following equations. 

\begin{equation}
\label{eq:2}
\begin{aligned}
    & i_t = \sigma(W_{xi} * X_t + W_{hi} * H_{t-1} + b_i) \\
    & f_t = \sigma(W_{xf} * X_t + W_{hf} * H_{t-1} + b_f) \\
    & C_t = f_t \circ C_{t-1} + tanh(W_{xc} * X_t + W_{hc} * H_{t-1} + b_c) \\
    & o_t = \sigma(W_{xo} * X_t + W_{ho} * H_{t-1} + b_o) \\
    & H_t = o_t \circ tanh(C_t)
\end{aligned}
\end{equation}

As previously $*$ denotes the convolution operation while $\circ$ denotes the Hadamard 
product. Furthermore, for timestep $t$ we find that $i_t$, $f_t$, $o_t$ are the outputs
of input gate, forget gate, and output gate respectively, $C_t$ is the cell output
and $H_t$ is the hidden state of a cell.

As explained with CNNs (\ref{S3.2.1}), ConvLSTM input sequence is usually ordered by 
closeness or adjacency in order to take advantage of the shared-weight scheme of the convolution.
Through this work, ConvLSTM will use the same input scheme as CNNs as presented in Fig.
\ref{fig:cnn_1}. Thus, convolution operations can make use of this type of spatial
relationship. 

\subsubsection{Graph Convolutional Neural Networks (GCN-LSTM)}
\label{S3.2.3}

While several proposals have been made during the last years to convolute over graphs,
we focus in a particular type presented in \cite{zhou_graph_2017} called High-Order and
Adaptive Graph Convolution, as it has shown good performance in a wide variety of problems.
In words of the authors, given a graph $\mathcal{G}$, the k-hop (k-th order) neighborhood
is defined as: $N_j = \{ v_i \in V |d(v_i, v_j ) \leq k \}$ for node $v_j$ . In fact,
the exact k-hop connectivity can be obtained by the multiplication of the adjacency
matrix $A$, giving as a result $A^{k}$. The convolution is defines as:

\begin{equation}
        \tilde{L}^{(K)}_{gconv,t} = (W_k \circ \tilde{A}^k)X_t + B_k,
\end{equation}

where $\circ$ refer to element-wise matrix product,
$B$ is the bias and $W$ a learnable matrix of weights.
$\tilde{A}^k$ is defined as $min\{A^{k} + I, 1\}$.

In order to adapt this kind of networks to spatio-temporal environments, a LSTM layer
is stacked with the convolutional one as with ConvLSTM in Section \ref{S3.2.2}.

Note that we use nodes to represent the spatial measurement
locations, which typically will be sensor stations or road segments,
and edges to represent the spatial segments connecting those sensing locations.
The adjacency matrix $A$ which defines these spatial segments is usually built based
on spatial metrics. For convenience and homogenization, we define for each dataset
$A$ as:

\begin{equation}
A_{i,j} = \begin{cases}
1 &\text{($i,j$ are neighbors)}\\
0 &\text{(otherwise)}
\end{cases}
\end{equation}

Where two locations $i$ and $j$ are considered neighbors if they are 
among the 4 closest areas without counting themselves. Through this definition of A, it is
trivial to see that the convolution over each space zone makes use of information based on
proximity.   

\subsection{Spatial agnosticism via Convolutional Networks}
\label{S3.3}

Now we show a series of spatial agnostic versions
based on the models presented through Section \ref{S3.2} for spatio-temporal regression. 
These methods will help us to test the main hypothesis of this work: whether introducing 
spatial-adjacency bias is unquestionably the best option or not. 
For this purpose, each agnostic version needs to fulfill two requirements:

\begin{itemize}
    \item No spatial information is introduced to the network.
    \item Past temporal information can be handled and introduced in the
      calculation of each new state.
\end{itemize}

By doing so, we will have several spatio-temporal methodologies that let
us contrast our main premise.

\subsubsection{Agnostic Convolutional Neural Network (A-CNN)}
\label{S3.3.1}

\begin{figure}[tbp]
\centering
\includegraphics[width=0.5\textwidth]{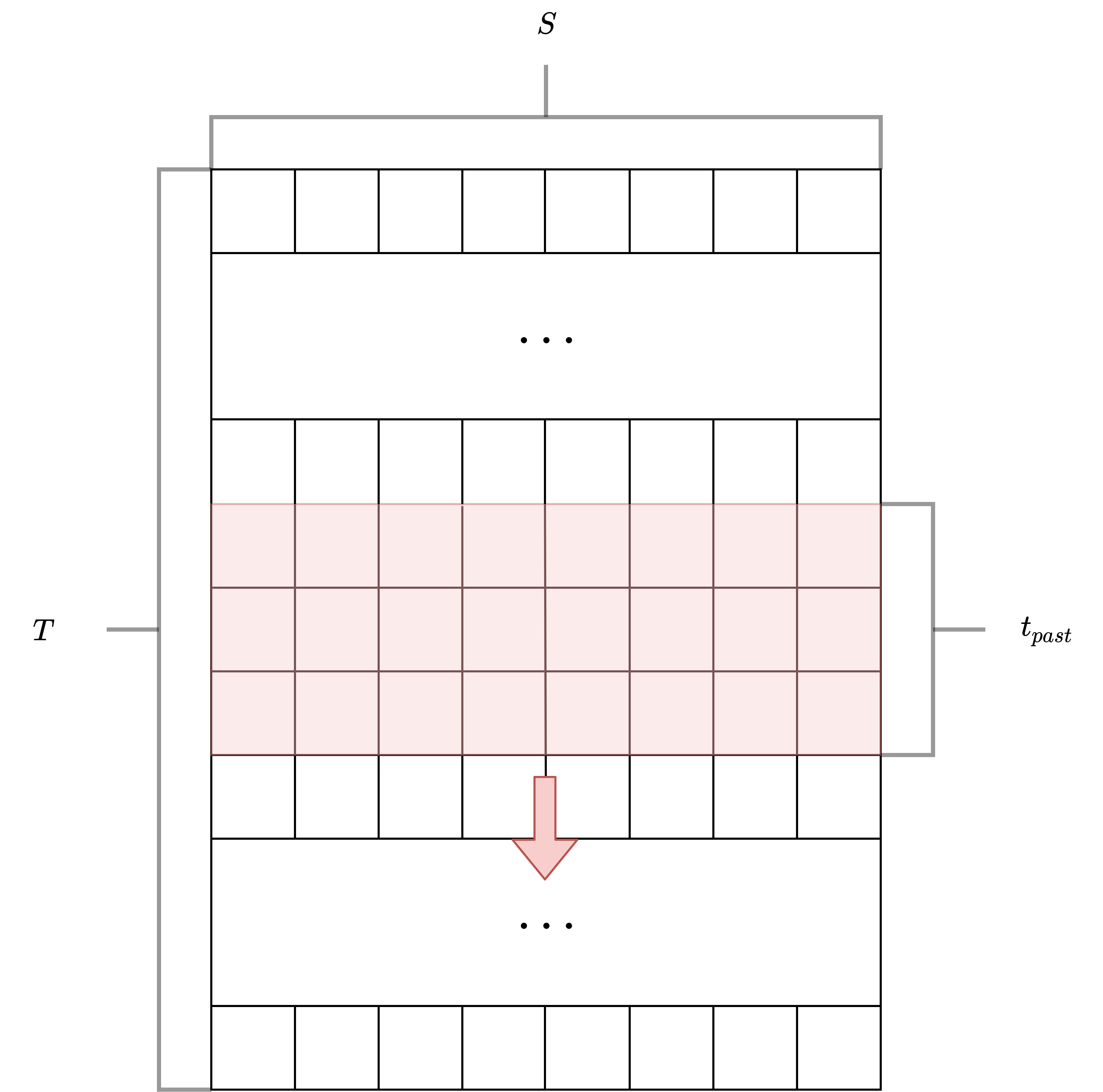}
\caption{Example of causal convolution spatially agnostic with $t_{past} = 3$ through a 
spatio-temporal sequence of just one variable as defined previously.}
\label{fig:sann_2}
\end{figure}

To define an agnostic version of CNNs, we can work from Equation \ref{eq:1}. However,
the kernel size is regularly used with equivalent values for its two dimensions 
$k_{1} = k_{2} = k$. In this case, not only this kernel uses different values for each 
component, but kernel size for spatial dimension must be equal to the number of spatial 
zones: $k_{2} = S$. As a result, the convolution operation is made over all locations at 
once. The kernel size in the temporal dimension is defined as $t_{past}$ and needs to be stipulated
as part of the network architecture. An example of this kind of filter can be found in 
Figure \ref{fig:sann_2}.

The temporal dimension is dominated by a causal convolution. Generally, causal convolution 
ensures that the state created at time $t$ derives only from inputs from time $t$ to $t - 
t_{past}$. In other words, it shifts the filter in the right temporal direction. Thus 
$t_{past}$ can be interpreted as how many lags are been considered when processing an 
specific timestep. Given that previous temporal states are taken into account for each step 
and that parameters are shared all over the convolution, this methodology might be seen as 
some kind of memory mechanism by itself. Unlike memory-based RNNs (like LSTMs and GRUs) where the 
memory mechanism is integrated solely by learned via the hidden state, in this case 
$t_{past}$ act as a variable that lets us take some control over this property.

\begin{figure}[tbp]
\centering
\includegraphics[width=0.5\textwidth]{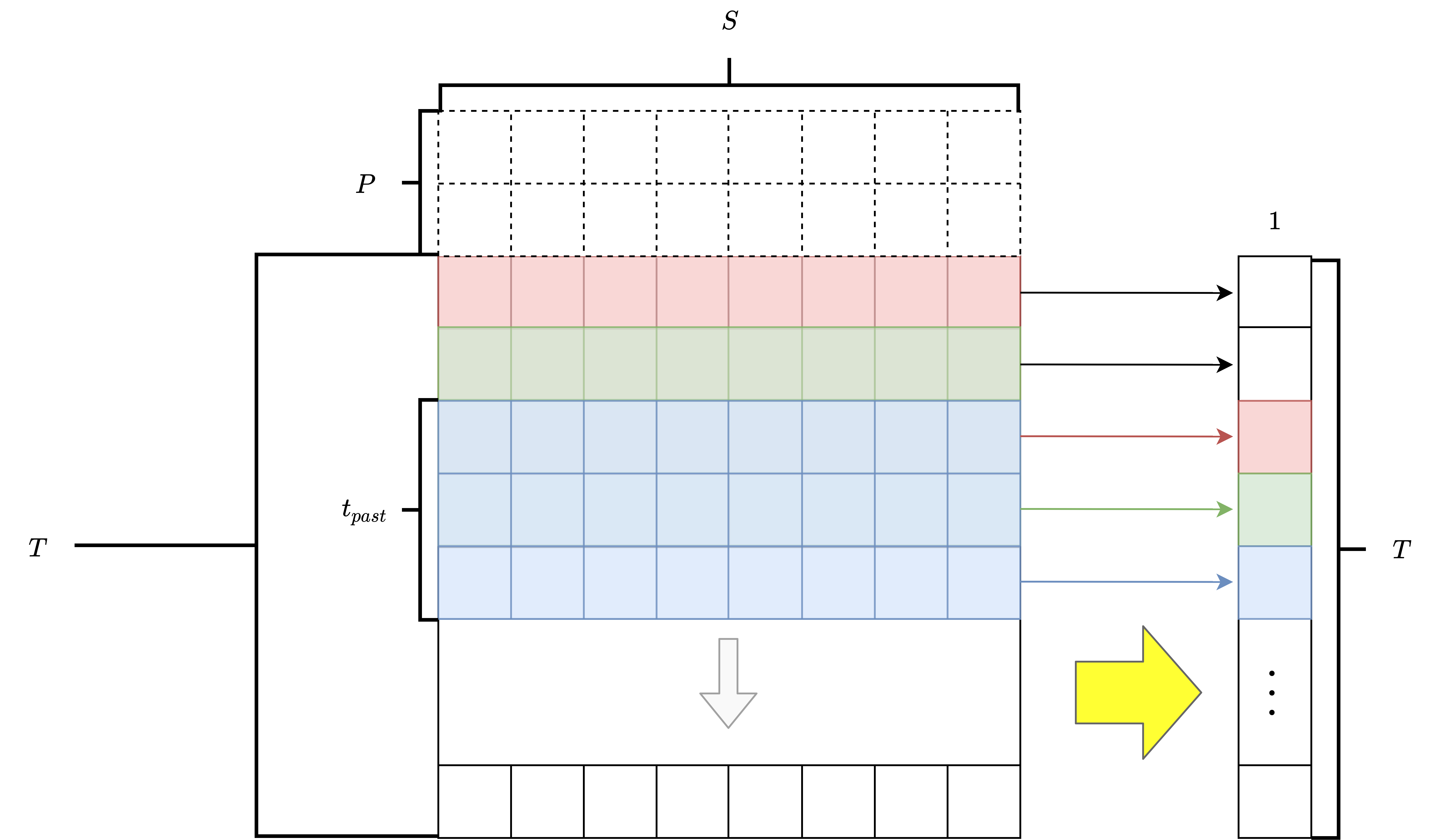}
\caption{Illustration of several step of the convolutional part of an agnostic convolutional
block. After moving all over the input sequence, a $T \times 1$ image is produced. This new
image compress information from all spatial locations and all input lags, keeping track of 
several of these ones in each convolution.}
\label{fig:sann_3}
\end{figure}

In order to ensure that each input timestep has a corresponding new state when
convolving, a padding of $P=t_{past}-1$ at the top of the input ``image'' is
required, and to guarantee temporal integrity, this padding must be done only at
the top. By using convolution in this form, once the kernel has moved over the
entire input image $T \times S$, the output image will be $T \times 1$. This
process is summarised in Fig \ref{fig:sann_3}.

\begin{figure}[tbp]
\centering
\includegraphics[width=0.5\textwidth]{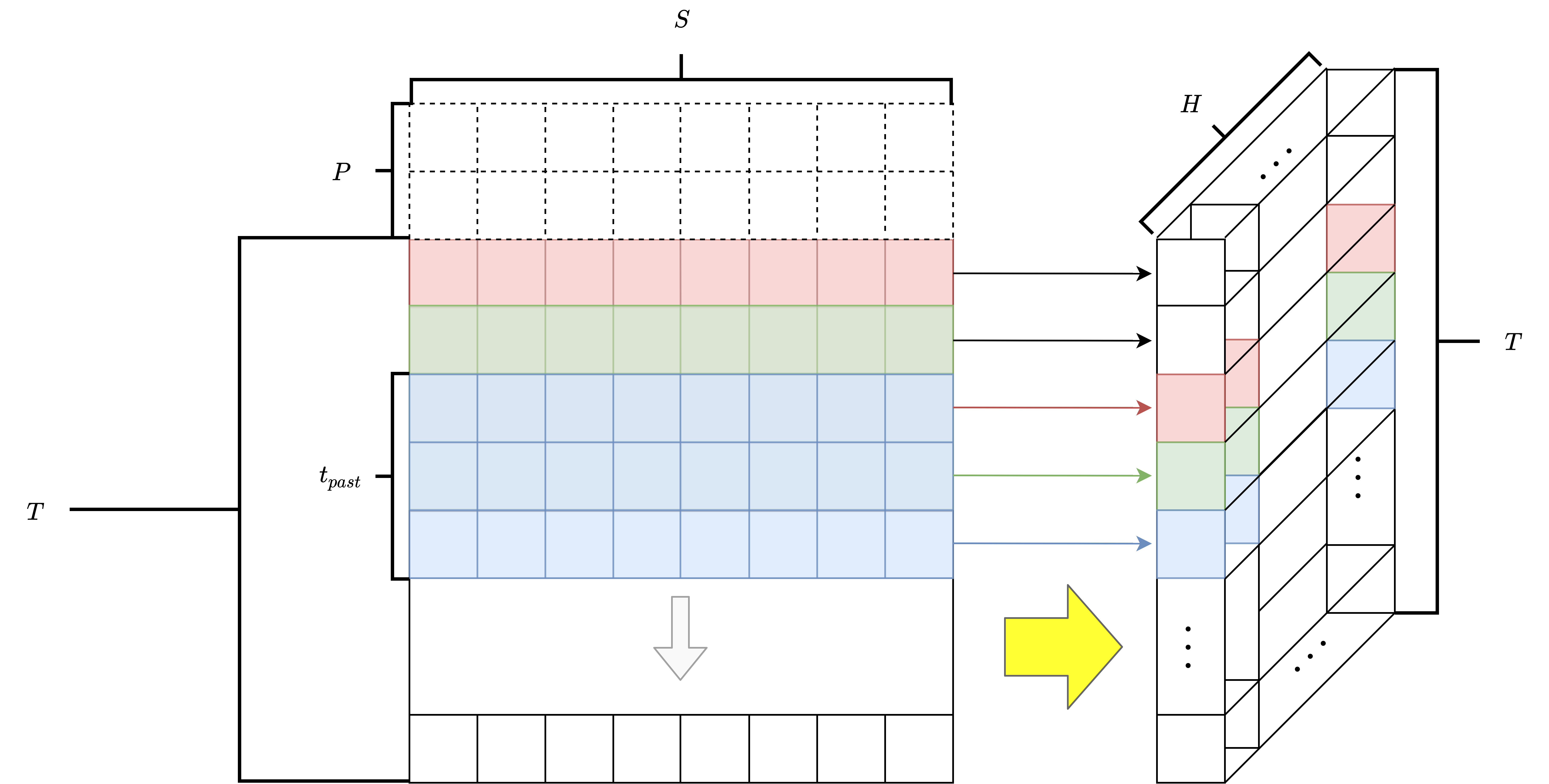}
\caption{By repeating operations described before, it is trivial to assemble hidden states as 
new channels in the latent sequence, meaning $T \times 1$ images with $H$ channels.}
\label{fig:sann_4}
\end{figure}

Now, if we repeat this operation $H$ times, we will create a new hidden state with $H$ channels 
an output an image with $H \times T$ dimensions as the example in Fig \ref{fig:sann_4}.

\begin{figure}[tbp]
\centering
\includegraphics[width=0.5\textwidth]{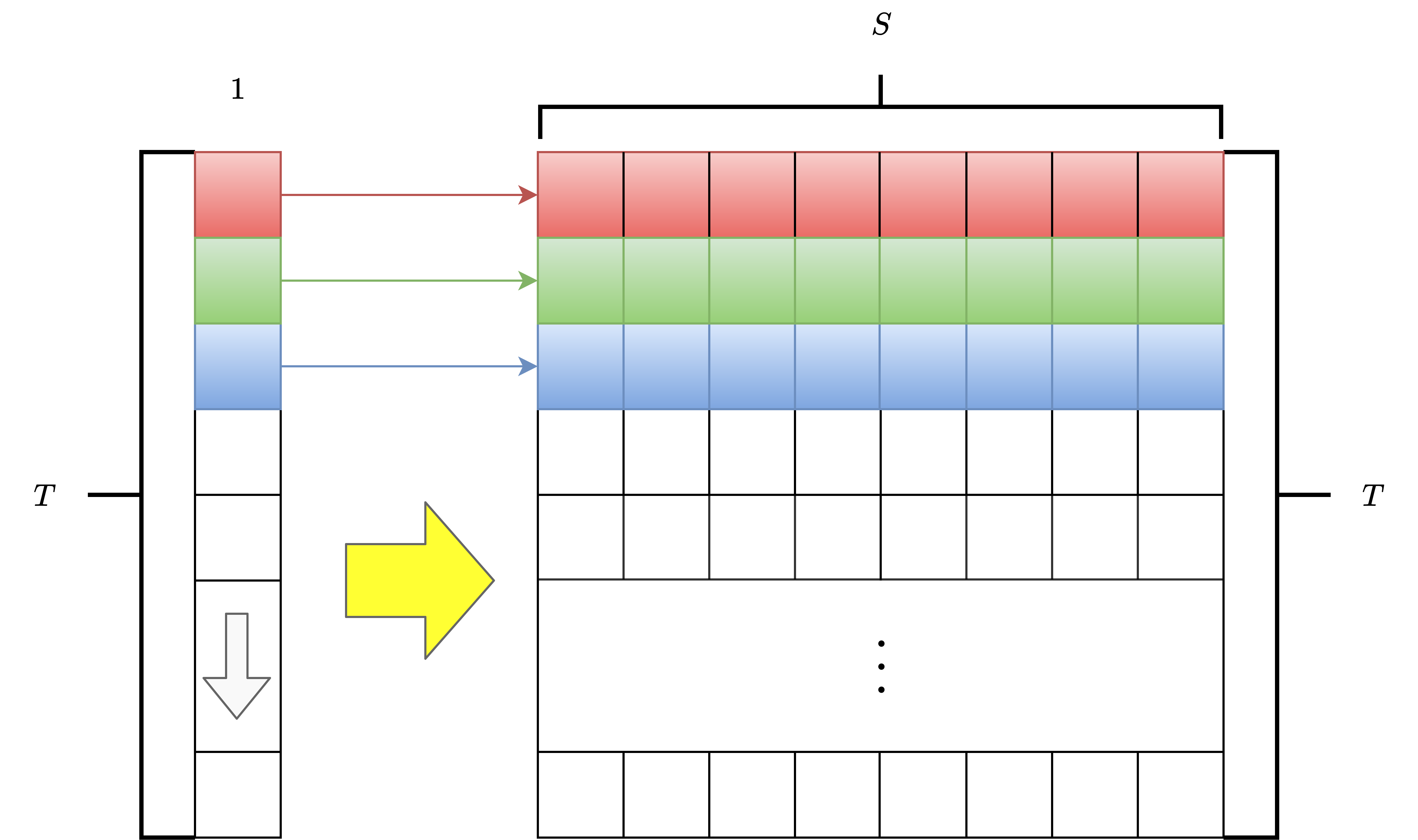}
\caption{Transposed convolution to produce a $T \times S$ images from $T \times 1$
latent sequences. Thanks to this process, we give the model same expressiveness
opportunities than traditional CNNs.}
\label{fig:sann_5}
\end{figure}

To give the network the opportunity to cover a spectrum of possibilities in terms of 
expressiveness as wide as a usual CNN for each channel, we simply use transposed 
convolution with a kernel size $k=(1,S)$ so the system can learn a $T \times S$ 
representation from a $T \times 1$ image. Fig \ref{fig:sann_5} illustrate this idea.

% \rnote{Igual lo de la transposed convolution entra muy forzado de repente}
% \jnote{Yo no lo veo forzado, la verdad. Lo que sí haría sería modificar el
%   tamaño de la figura (de todas) para que los textos que hay en ellas tengan
%   siempre un tamaño homogéneo.}

\begin{figure*}[btp]
\centering
\includegraphics[width=1\textwidth]{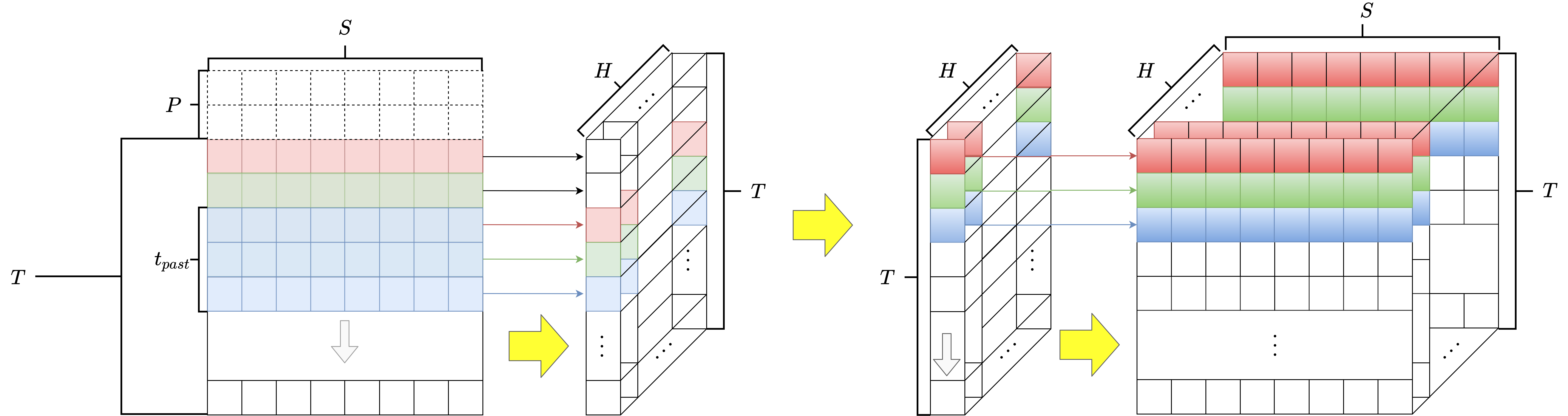}
\caption{Representation of a complete agnostic convolutional block. By assembling operations
described before, from a $T \times S$ it is trivial to create hidden states capable of representing 
equivalent expressions compared to a traditional CNN, meaning $T \times S$ images with $H$ channels.}
\label{fig:sann_6}
\end{figure*}

Evidently, our new representation is usually composed by $H$ hidden states, so this 
transposed convolution will use $H$ filters. Finally, the complete procedure for an entire 
agnostic convolutional block is described graphically in Fig \ref{fig:sann_6}.

Obviously, there are no restrictions with respect to the width dimension. For simplicity, 
we have considered convolutions in which only the number of channels is changed, meaning
that images keep an $T \times S$ structure during all the computations. As we have described
previously, this will help to normalize our experiments. However, as with CNNs, 
the dimensionality of hidden and output states might be different. Over all this process,
the arrangement of the spatial dimension (columns) has no effect, meaning that it is not
it is not necessary to map the study areas in any specific way with the input of the
network, as was done with the CNNs (Fig. \ref{fig:cnn_1}).

\subsubsection{Agnostic ConvLSTM (A-ConvLSTM)}
\label{S3.3.2}

Once that the agnostic procedure for convolving has been presented in the previous section,
the A-ConvLSTM is governed by Equation \ref{eq:2} but changing the traditional convolution
for this new approach. The only difference with respect to A-CNN is that there is no need
of causal convolution over the temporal dimension, as the LSTM module can handle it.
Therefore, the input sequence lacks any spatial ordering procedure, letting us
define this dimension arbitrarily.

\subsubsection{Agnostic Graph Convolutional Network (A-GCN-LSTM)}
\label{S3.3.3}

Following the structure of the GCN-LSTM presented in Section \ref{S3.2.3},
its agnostic version is simply to define the adjacency matrix as the identity matrix:
$A = I_{S}$. Thus, we can make sure that no spatial relation is being introduced
or modeled explicitly. Otherwise, the network has the same functioning and
characteristics as described before.

A comparison between each traditional model and its agnostic version is shown
in Table \ref{tab:sum_sp}. There it is summarized how each model formalize \textit{prior}
information about the spatial dimension and how it affects their use.

\begin{table}[btp]
  \centering
  \caption{Summary of spatial treatment of each model.}
  \label{tab:sum_sp}
  \scalebox{0.85}{
  \begin{tabular}{lll}
    \toprule
        &  Traditional & Agnostic \\ \midrule
    CNN & \begin{tabular}[c]{@{}l@{}}Shared kernel among locations. \\ Ordering of spatial dimension \\ based on closeness\end{tabular} &  \begin{tabular}[c]{@{}l@{}}One kernel for each location.\\ No ordering needed\end{tabular}   \\ \\
    ConvLSTM & \begin{tabular}[c]{@{}l@{}}Shared kernel among locations. \\ Ordering of spatial dimension \\ based on closeness\end{tabular} & \begin{tabular}[c]{@{}l@{}}One kernel for each location.\\ No ordering needed\end{tabular}  \\ \\
    GCN-LSTM & \begin{tabular}[c]{@{}l@{}}Adjacency matrix defined \\ by proximity\end{tabular} &  \begin{tabular}[c]{@{}l@{}}Identity matrix as \\ adjacency matrix\end{tabular} \\ \bottomrule
  \end{tabular}
  }
\end{table}

% \begin{table}[btp]
%   \centering
%   \caption{Summary of spatial treatment of each model.}
%   \label{tab:sum_sp}
%   \scalebox{0.85}{
%   \begin{tabular}{lccc}
%     \toprule
%         &  CNN & ConvLSTM & GCN-LSTM \\ \midrule
%     Traditional & \begin{tabular}[c]{@{}l@{}}Shared kernel between locations. \\ Ordering of spatial dimension \\ based on closeness\end{tabular} &  \begin{tabular}[c]{@{}l@{}}Shared kernel between locations. \\ Ordering of spatial dimension \\ based on closeness\end{tabular} & Adjacency matrix defined by proximity  \\ 
%     Agnostic & \begin{tabular}[c]{@{}l@{}}One kernel for each location.\\ No ordering need\end{tabular} & \begin{tabular}[c]{@{}l@{}}One kernel for each location.\\ No ordering need\end{tabular} & Identity matrix as adjacency matrix \\  \bottomrule
%   \end{tabular}
%   }
% \end{table}

\subsection{Regressor block}
\label{S3.4}

Through Sections \ref{S3.2} and \ref{S3.3}, we have explored how to use convolution
operations to learn a new hidden representation of the input sequence as an image
with and without using \textit{prior} spatial information or closeness assumptions.
Now, in order to make a fair comparison between traditional networks and their
respective agnostic versions, we have to carefully use this latent representation with
$T \times S \times H$ dimensions (common to all models presented) to get
a new $T' \times S$ predicted image. While this process can be done in multiple
ways, it is desirable for this regressor block to fulfill several conditions:

\begin{enumerate}[label=(\arabic*)]
    \item The same strategy has to be applicable to all models studied in this work.
    
    \item It can not explicitly share information between elements of the spatial dimension. This way, we make sure that space is only treated in the convolutional block of each model and our results are not contaminated from other parts of the network.
    
    \item The number of parameters needs to be as low as possible and space-independent. Thus, we avoid overfitting or overinfluence problems. 
    
    \item Lastly, although we have not found an option that is completely network 
    architecture-independent (you can get a similar size of hidden dimension or total
    number of parameters, but not both), it is highly desirable that this regressor
    layer does not undergo too much variability between models.
\end{enumerate}

\begin{figure}[tbp]
\centering
\includegraphics[width=0.5\textwidth]{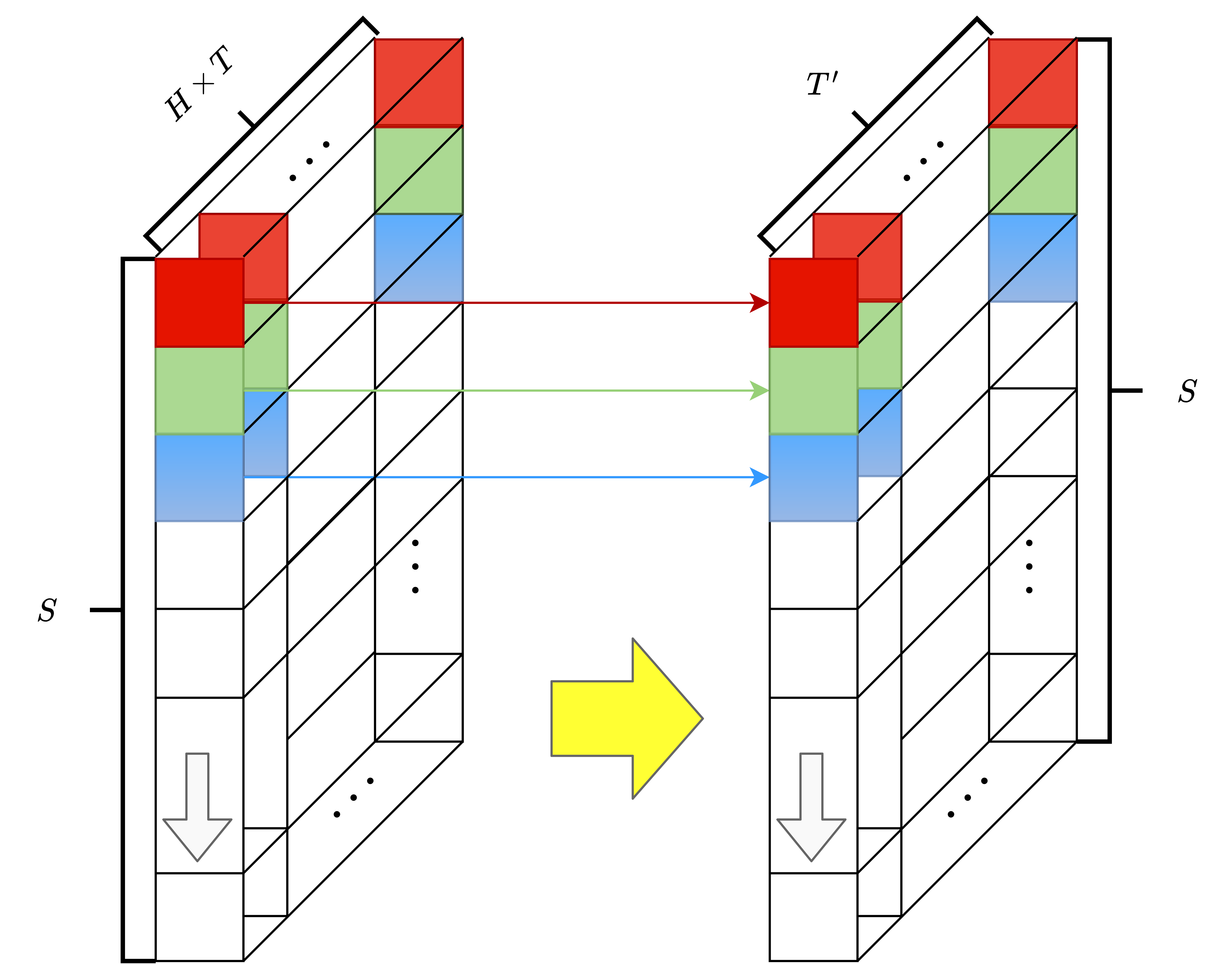}
\caption{Regressor block that satisfy applicability, spatial agnosticism and simplicity. By using a 1D convolution over the latent image $H \times T \times S$, we can produce a $T' \times S$ sequence that correspond to our forecast.}
\label{fig:sann_7}
\end{figure}

A naive and simple approach would be using 1D convolutions after reshaping the
$H \times T \times S$ image into a $(H \cdot T) \times S$, with $H \cdot T$
being the number of input channels. By convolving trough the spatial dimension
with a kernel size of $k = 1$ and an output number of channels of $T'$, we can
be sure no information is shared through this dimension (2) and the number of
parameters, which is $H \cdot T \cdot T'$, remains low compared to the complete
network (3). Furthermore, all models that we will compare are based on a
convolutional block which outputs an $H \times T \times S$, meaning that this
regressor scheme can be applied to all of them, helping to standardize our
experiments (1). As $T$ is the same for all models and $H$ never diverges more
than one order of magnitude, we can be sure this layer has a similar impact for
all cases (4). Fig \ref{fig:sann_7} summarize this block.

Although other options have been considered, as 2D convolutions and dense
layers, they fail to meet some conditions or need fine-tuning for each problem
and model, making them less suitable for a fair comparison.

% \rnote{Ha habido una auténtica batalla intentando buscar la mejor forma de
%   conseguir esto para que sea lo más justo y similar posible sin importar el
%   modelo y el problema. No sé si tiene cabida o no.} \jnote{Yo creo que sí
%   podrías profundizar un poco más, si quisieras dar cuenta de esa ``batalla'',
%   pero no es imprescindible.}

\subsection{Temporal vs spatial distribution}
\label{S3.5}

Our work is based on the hypothesis that real spatio-temporal series might not
share a similar behavior in their two dimensions. Even the well known fact that
closer, spatially speaking, locations behaves similarly does not always suit
well, meaning that the parameter sharing scheme of traditional CNNs might not be
the best option. Concretely, when dealing with real problems, the system might
have a high dependency on non-spatial phenomena and data collection can have a
great impact. As a result, closeness information can be lost or modified.

On the contrary, temporal information (or distribution) usually keeps the same
structure for a wide range of problems. As air quality and mobility are high
correlated to human being, the temporal pattern of this kind of series for each
location tends to remain alike.

In order to prove our hypotheses, we will make use of statistical tools that
characterize the aforementioned information.

\subsubsection{Spatial dimension: Moran's $I$}
\label{S3.3.1}

According to \cite{Lee2017}, "Spatial autocorrelation or spatial dependence can
be defined as a particular relationship between the spatial proximity among
observational units and the numeric similarity among their values; positive
spatial autocorrelation refers to situations in which the nearer the
observational units, the more similar their values (and vice versa for its
negative counterpart)... This feature violates the assumption of independent
observations upon which many standard statistical treatments are predicated."
This property, which is precisely what we are interested in, can be measured by
the well know Moran's $I$. This test will let us quantify the degree of spatial
autocorrelation existing in the different datasets that we will use between
close locations taking into account this interdependency. As it is a test,
Moran's $I$ comes with a p-value which typifies statistical significance of the
result. It is defined as:

\begin{equation}
    \label{eq:3}
    I = \frac{S}{W} \frac{\sum_{i} \sum_{j} w_{ij} (x_{i} - \bar{x})(x_{j} - \hat{x})}{\sum_{i}(x_{i} - \bar{x})^{2}}
\end{equation}

where $S$ is the number of spatial units indexed by $i$ and $j$, $x$ is the
variable of interest, $\bar {x}$ is the mean of $x$, $w_{ij}$ is a matrix of
spatial weights based on neighbors, and $W$ is the sum of all $w_{ij}$. As its
value varies usually between $-1$ and $+1$, it is easily interpretable.
Concretely, $+1$ implies similar values for close locations, $0$ a random
arrangement, and $-1$ opposite values.

As we also have a temporal dimension, we will average $I$ for all timesteps.
Through this test we want to compute solely spatial autocorrelation, without
intervention of temporal relations between locations.

\subsubsection{Temporal dimension: Adaptative Temporal Dissimilarity Measure}
\label{S3.3.2}

To compare the similarity between different time series (in our case, different
spatial points) the same problem arises than with spatial autocorrelation: due
to the interdependence relationship between measurements classical correlation
index can not be applied. For example, Euclidean, Fréchet distances and Dynamic
time warping are well known and widely used techniques when measuring time
series similarity but do not handle the aforementioned issue well. To solve this
problem, \cite{chouakria_adaptive_2007} proposed the Adaptative Temporal
Dissimilarity Measure (ATDM) as an index that lets us measure the similarity
between time series more robustly as it balances the proximity with respect to
values and the proximity with respect to behavior. Ir writes as:

\begin{equation}
\begin{gathered}
    \label{eq:4}
    \mathrm{ATDM}(X_{T,i},X_{T,j}) = \\ f(\mathrm{cort}(X_{t,i},X_{t,j})) \cdot \delta(X_{t,i},X_{t,j}),
    \end{gathered}
\end{equation}

where $\delta$ references a classical distance (we will use Euclidean) and $\mathrm{cort}$ is

\begin{equation}
\begin{gathered}
    \label{eq:5}
    \mathrm{cort}(X_{T,i},X_{T,j}) = \\ \frac{\sum^{T-1}_{t} (X_{t+1,i} - X_{t,i}) (X_{t+1,j} - X_{t,j})}{\sqrt{\sum^{T-1}_{t} (X_{t+1,i} - X_{t,i})^{2}} \sqrt{\sum^{T-1}_{t}(X_{t+1,j} - X_{t,j})^{2}}}.
\end{gathered}
\end{equation}

Lastly, $f$ writes as follow:

\begin{equation}
    \label{eq:6}
    f(x) = \frac{2}{1+\exp{(kx)}}, k \geq 0. 
\end{equation}

With this metric, the distance is squeezed into a coefficient in the interval $(0,2)$. When the 
correlation coefficient is 0, the ATDM is 1, and the correlation is not significant. When the 
correlation is positive, the value of the ATDM is less than 1; the more similar the two series are, 
the smaller the value is. On the contrary, the ATDM is more than 1 if the correlation is negative. The
less similar the two series are, the larger the value is.

Thus, we can average the ATDM between all locations pairs for each spatio-temporal series. As this 
measure takes into account both values and behavior of the series, we can approximately get a global 
measure of temporal distribution similarity among points for each dataset. 

When working with real data, in which depending on time granularity local properties of time series 
might be noisy, ATDM might not extract information correctly. In order to solve this, we compute an 
adjusted ATDM coefficient (ATDM$_{adj}$) which uses a smoother version of the input series as we are 
interested in global behavior of the temporal distribution. Concretely, we use moving average as it 
is simple and has shown to be a good approximator for time series. As moving average just smooth the 
series, we do not expect to corrupt the coefficient between series which are not really temporally 
correlated.

% %%%%%%%%%%%%%%%%%%% %
% EXPERIMENTAL DESIGN %
% %%%%%%%%%%%%%%%%%%% %
\section{Experimental design}
\label{S4}

\subsection{Data description}
\label{S4.1}

The different forecasting problems and the corresponding datasets are described below. Main dataset characteristics and statistics are provided in Table \ref{tab:statistics}.

\begin{itemize}
\item \textbf{AcPol dataset:} Provided by the Municipality of Madrid through its
    open data portal\footnote{\textit{Portal de datos abiertos del Ayuntamiento de
    Madrid}: \url{https://datos.madrid.es/portal/site/egob/}\label{note1}}. Acoustic 
    pollution in Madrid in decibels, it measures equivalent continuous level with A frequency weighting, which is the assumed noise level constant and continuous over a period of time, corresponding to the same amount of energy than that actual variable level measured in the same period.

\item \textbf{Beijing dataset:} Presented by \cite{bbliaojqZhangKDD18deep}, it consist of 
    traffic speed measurements for $~15000$ road segments recorded per minute. To make the 
    traffic speed predictable for each road segment, it is aggregated via moving average in 
    15 minutes intervals. For this work, we select a subgroup of road segments spatially 
    close.

\item \textbf{BiciMad dataset:} Supplied by EMT (Municipal Transport Company for its
    initials in Spanish) through its
    open data portal\footnote{\textit{Portal de datos abiertos EMT}: 
    \url{https://opendata.emtmadrid.es/Datos-estaticos/Datos-generales-(1)}\label{note2}}. In this case
    we tackle the bike sharing demand prediction by aggregating the overall number of bikes per station
    and timestep.

\item \textbf{LOOP dataset:} It contains data collected from inductive traffic loop detectors 
    deployed on four connected freeways (I-5, I-405, I-90, and SR-520) in the Greater     
    Seattle Area. It can be found in \cite{cui_traffic_2019}.

\item \textbf{MATRA dataset:} This dataset contains historical data of traffic measurements
    in the city of Madrid. The measurements are taken every 15 minutes at each point, 
    including traffic intensity in number of cars per hour. Data is aggregated for each 
    hour. While a dense and populated network of over 4.000 sensors is available, we decided
    to simplify and use only a selection of them. Available in the Municipality of Madrid
    open data portal$^{\ref{note1}}$.

\item \textbf{METR-LA dataset:} This dataset contains traffic information recopilated from 
    loop detectors in the highway of Los Angeles County. We use the partition provided by \cite{li_diffusion_2018}.

\item \textbf{NO2 dataset:} $NO_{2}$ in the city of Madrid. Hourly data for all measurement
    stations which include this pollutant. Available in the Municipality of Madrid open data portal$^{\ref{note1}}$.

\item \textbf{NYTaxi dataset:} Provided by Taxi $\&$ Limousine Commission 
    \footnote{\textit{NYCTaxi and Limousine Commission (TLC) Trip Record Data}:       
    \url{https://www1.nyc.gov/site/tlc/about/tlc-trip-record-data.page}\label{note3}},
    it consist of taxi trip location and duration in the city of New York. We focus our work
    in forecasting number of taxi travels for each New York neighborhood with an average minimum number of one trip per day.

\item \textbf{O3 dataset:} $O_{3}$ in the city of Madrid. Hourly data for all measurement
    stations which include this pollutant. Available in the Municipality of Madrid open data portal$^{\ref{note1}}$.

\item \textbf{PEMS-BAY dataset:} This traffic dataset is collected by California
    Transportation Agencies (CalTrans) Performance Measurement System (PeMS). We use the partition provided by \cite{li_diffusion_2018}.

\end{itemize}

All datasets are Z-Score normalized by spatial point. We take as reference previous work as 
a criterion to choose $T$ and $T'$. Thus, we can be sure of the plausibility of the results 
for all models. When no previous work is known, we use autocorrelation as a measurement of 
number of minimum lags ($T$) and focus only on a single timestep prediction ($T'=1$).

From Table \ref{tab:statistics} we can see how our chosen datasets cover a wide range of 
spatio-temporal circumstances and the high variety and variability of data. Also, our main 
hypotheses are confirmed: Moran's $I$ show a clear no-spatial autocorrelation pattern for our series, 
and although not completely uncorrelated, most series are close to $0$. All p-values are lower than 
$0.05$. It is worth noting as proof of plausibility for these values that \cite{lu_modeling_2018} 
computed the coefficient $I$ for the complete Beijing traffic dataset at some hours, reporting a 
similar value to ours. ATDM values tend to be low, which is representative of similar temporal 
distributions in the datasets. As we expected, ATDM$_{adj}$ represents better this idea. Datasets with
a clear temporal pattern but locally noisy, as Beijing, LOOP, and PEMS-BAY, are better described by 
this coefficient.

Given that spatial locations are by default in arbitrary order, it is necessary to sort and structure 
them in order to fully exploit spatial information with traditional models.
By computing a hierarchical tree (dendrogram) using an agglomerative hierarchical clustering algorithm
and traversing recursively the tree it is possible to approximately sort the points by distance.

\begin{table*}[btp]
\centering
\caption{Details of data through experiments. $Dates$ reflects starting and ending points of data, $Timestep$ corresponds to the duration of one timestep. $T$, $T'$ and $S$ were defined in Section \ref{S3.1} as input timesteps, output timesteps and number of spatial locations. $Mean$, $Median$, and $Std$ condense main data statistics. $ATDM$, $ATDM_{adj}$, and $Moran's \: I$ summarize information about spatial and temporal distribution similarity between locations.}
\label{tab:statistics}
\scalebox{0.85}{
\begin{tabular}{lcllllllllll}
  \toprule
  Dataset & Dates & Timestep & $T$ & $T'$ & $S$ & Mean & Median & Std & ATDM & ATDM$_{adj}$ & Moran's $I$ \\
  \midrule
  AcPol & 2014/01/01 -- 2019/03/31 & 1 day & 7 & 1 & 30 & 56.8 & 60.2 & 15.1 & 0.36 & 0.36 & 0.03 \\
  Beijing & 2017/01/04 -- 2017/05/31 & 15 min & 10 & 1 & 200 & 29.0 & 28.7 & 9.3 & 0.69 & 0.27 & 0.20 \\
  BiciMad & 2019/01/01 -- 2019/06/30 & 1 hour & 6 & 1 & 168 & 0 & 0 & 3.2 & 1.04 & 1.03 & 0.12 \\
  LOOP & 2015/01/01 -- 2015/03/31 & 5 min & 10 & 1 & 323 & 57.2 & 60.6 & 11.8 & 0.84 & 0.47 & 0.31 \\
  MATR & 2018/01/01 -- 2019/12/31  & 1 hour & 24 & 6 & 120 & 445.5 & 254.8 & 539.6 & 5.6E-4 & 4.8E-4 & 0.09 \\
  METR-LA & 2012/03/01 -- 2012/06/30 & 5 min & 12 & 3 & 207 & 53.4 & 62.3 & 20.6 & 0.02 & 0.02 & 0.24 \\
  NO2 & 2017/01/01 -- 2019/12/31 & 1 hour & 48 & 48 & 24 & 37.5 & 29 & 28.9 & 0.04 & 0.0 & 0.13 \\
  NYTaxi & 2016/01/01 -- 2016/06/30  & 1 hour & 6 & 1 & 70 & 4.8 & 0 & 11.3 & 0.55 & 0.34 & 0.24 \\
  O3 & 2017/01/01 -- 2019/12/31 & 1 hour & 48 & 48 & 14 & 50.6 & 50 & 34.3 & 0.03 & 0.0 & 0.11 \\
  PEMS-BAY & 2017/01/01 -- 2017/05/31 & 5 min & 12 & 3 & 325 & 62.6 & 65.3 & 9.6 & 0.64 & 0.15 & 0.23 \\
   \bottomrule
\end{tabular}
}
\end{table*}

\subsection{Architecture models}
\label{S4.2}

We compare agnostic models with widely used spatio-temporal series regression models 
based on the convolution operator. Details concerning its architectures are:

\begin{itemize}

    \item \textbf{A-CNN:} Through the process batch normalization
    and ReLU activation function are used.
    
    \item \textbf{CNN:} A standard CNN followed by a batch normalization layer
    and ReLU activation function. It uses a 3$\times$3 kernel.
    
    \item \textbf{A-ConvLSTM:} ReLU activation function after convolution. No batch
    normalization.
    
    \item \textbf{ConvLSTM:} A standard ConvLSTM that uses a
    3$\times$3 kernel. ReLU activation function after convolution. No batch
    normalization.
    
    \item \textbf{A-GCN-LSTM:} ReLU activation function.
    
    \item \textbf{GCN-LSTM:} A classical approach for GCN which let us 
    exploit explicitly information from the $k$-hop ($k$-th order) 
    neighborhood of each node in the graph. In our experiments, we set $k=3$ and
    use ReLU activation function.
    
\end{itemize}

As we are interested in deepening in how the convolution operator and the spatial dimension are related, we do not include any RNN or FNN based approach.

\subsection{Experimental design}
\label{S4.3}

In order to make a comparison as fair as possible, we decided to proceed with
all models as follows:

\begin{itemize}
\item They will consist uniquely in a convolutional layer and a regressor layer.
  For all of them, the convolutional layer will enrich input information by
  constructing a $H \times T \times S$ image from a $T \times S$ sequence as
  described in Section \ref{S3.1}.
\item The regressor layer consists of a 1D convolution, as explained in Section
  \ref{S3.4}. Thus, we make sure no model is taking advantage or exploiting
  further spatial information.
\item The number of parameters in the convolutional layer need to remain similar
  and in the same magnitude order. Given regressor layer's architecture and the
  fact that it is the same for all models, we expect that this is enough to
  eliminate possible bias.
\item A weight decay (L2 regularization) of $10^{-3}$ is used to prevent
  overfitting.
\end{itemize}

Some other minor details are that all the models are trained using the mean
squared error (MSE) as objective function with the RMSprop optimizer, as it has
shown good performance in non-stationary scenarios. Batch size is 256, momentum
is set to 0.9, the initial learning rate is 0.001 and both early stopping and
learning rate decay are implemented in order to avoid overfitting and improve
performance. The experiments are run in a NVIDIA RTX 2070.

As we have standardized the experiments, no hyperparameter tuning is needed in
general. Solely $t_{past}$ for A-CNN needs to be adjusted, which will be tuned
via standard grid search.

\subsection{Validation and error metrics}
\label{S4.4}

As stated in \cite{bergmeir_note_2018}, standard $k$-cross-validation is the way
to go when validating neural networks for time series if several conditions are
met. Specifically, that we are modeling a stationary nonlinear process, that we
can ensure that the leave-one-out estimation is a consistent estimator for our
predictions and that we have serially uncorrelated errors.

While the first and the third conditions are trivially fulfilled for our
problem, the second one needs to be specifically treated. Given that some input
sequences might share elements among different sets(training, validation and
test), \textit{prior} information could be entangled leading to data leakage.
Due to this problem, it is not possible to create random folds and it is
necessary to specify a separation border among previously defined sets.
Particularly, a 10-cross-validation scheme without repetition is used during all
experiments, with a 80\%/10\%/10\% scheme for train/validation/ test sets for
each fold. 

To evaluate the precision of each model, we computed root mean squared error
(RMSE) and bias. In a
spatio-temporal context \cite{wikle_spatio-temporal_2019}, they are defined as:
\begin{equation}
  \mathrm{RMSE} = \sqrt{\frac{1}{T'S} \sum_{i = 1}^{T'} \sum_{i = 1}^{S} (\tilde{x}_{t'_i,s_j} - x_{t'_i,s_j})^{2}}, \label{eq:7}
\end{equation}
\begin{equation}
  \mathrm{bias} = \frac{1}{T'S} \sum_{i = 1}^{T'} \sum_{j = 1}^{S} ( \tilde{x}_{t'_i,s_j} - x_{t'_i,s_j} ), \label{eq:8}
\end{equation}

For all these metrics, the closer to zero they are the better the performance
is. While RMSE already provides a dispersion measure respect to real series,
bias is better to find particular predispositions when making predictions. 

% %%%%%%%%%%%% %
% RESULTS %
% %%%%%%%%%%%% %
\section{Results}
\label{S5}

\subsection{Performance comparison}
\label{S5.1}

A general comparison of the different error metrics can be seen in Table
\ref{tab:res_total}.

\begin{table*}[btp]
  \centering
  \caption{Average performance per model and dataset. For a more detailed view
    of error metrics distribution, see Fig. \ref{fig:res_1}.}
  \label{tab:res_total}
  \resizebox{0.75\textwidth}{!} {
    \begin{tabular}{@{}lllllllllll@{}}
      \toprule
          & \multicolumn{2}{l}{AcPol} & \multicolumn{2}{l}{Beijing} & \multicolumn{2}{l}{BiciMad} & \multicolumn{2}{l}{LOOP} & \multicolumn{2}{l}{MATR} \\ 
      \cmidrule(l){2-3}
      \cmidrule(l){4-5}
      \cmidrule(l){6-7}
      \cmidrule(l){8-9}
      \cmidrule(l){10-11}
      
            & RMSE     & Bias     & RMSE     & Bias    & RMSE     & Bias      & RMSE      & Bias     & RMSE      & Bias   \\ \midrule
A-CNN       & 7.06     &  0.22    & 2.74     & 0.02    & 2.75     & -2.0E-4   & 5.06      & -0.07    & 115.65    & -5.30  \\
CNN         & 8.52     & -0.04    & 4.52     & -0.02   & 2.94     & -0.01     & 4.59      & 0.05     & 141.99    & 0.13   \\ \midrule
A-ConvLSTM  & 6.22     & 0.04     & 2.64     & 7.3E-3  & 2.74     & -8.3E-3   & 4.52      & 0.02     & 111.74    & -0.06  \\
ConvLSTM    & 5.46     & -3.0E-3  & 2.26     & -0.15   & 2.89     & -0.01     & 3.71      & 0.03     & 115.29    & -2.85  \\ \midrule
A-GCN-LSTM  & 7.45     & 0.17     & 2.88     & 0.02    & 2.76     & 6.7E-4    & 5.77      & -0.53    & 136.48    & 2.97   \\
GCN-LSTM    & 8.01     & 0.03     & 2.76     & 0.09    & 2.70     & 6.2E-3    & 5.02      & -0.18    & 132.14    & 0.43   \\ \bottomrule
    \end{tabular}
  }
  % \end{table}
  \\\vspace{0.1cm}
  % \begin{table}[]
  \resizebox{0.75\textwidth}{!} {
    \begin{tabular}{@{}lllllllllllllll@{}}
      \toprule
          & \multicolumn{2}{l}{METR-LA} & \multicolumn{2}{l}{NO2} & \multicolumn{2}{l}{NYTaxi} & \multicolumn{2}{l}{O3} & \multicolumn{2}{l}{PEMS-BAY} \\ 
          \cmidrule(l){2-3}
      \cmidrule(l){4-5}
      \cmidrule(l){6-7}
      \cmidrule(l){8-9}
      \cmidrule(l){10-11}
       
            & RMSE     & Bias     & RMSE     & Bias     & RMSE     & Bias     & RMSE      & Bias    & RMSE     & Bias   \\ \midrule
A-CNN       & 9.52     &  0.29    & 23.17    & 0.27     & 2.97     & 0.02     & 22.13     & 1.51    & 3.98     & -0.03  \\
CNN         & 10.00    & -0.01    & 24.26    & -0.03    & 3.52     & 0.06     & 23.30     & 0.607   & 3.98     & 0.03   \\ \midrule
A-ConvLSTM  & 9.13     & 0.24     & 22.79    & 0.01     & 2.86     & 1.3E-4   & 21.56     & 1.00    & 3.66     & -0.03  \\
ConvLSTM    & 7.86     & -0.05    & 24.25    & 0.31     & 3.16     & -0.01    & 21.56     & 0.71    & 2.42     & 0.04   \\ \midrule
A-GCN-LSTM  & 10.14    & 0.14     & 23.51    & 0.23     & 2.87     & 2.7E-3   & 22.71     & 0.25    & 4.16     & 0.14   \\
GCN-LSTM    & 10.17    & -0.58    & 24.98    & -0.90    & 2.88     & 0.01     & 23.39     & 1.75    & 4.07     & -0.68  \\ \bottomrule
    \end{tabular}
  }
\end{table*}

% \begin{table}[btp]
%   \centering
%   \caption{Average error metrics for each model and ranking depending on RMSE
%     value.}
%     % \jnote{Agregar RMSEs de magnitudes distintas no es muy correcto, ¿no?
%     %   ¿No habría que ponderar por el rango (o la desviación típica) de cada variable?}}
%   \label{tab:res_average}
%   \scalebox{0.85}{
%     \begin{tabular}{lcccc}
%       \toprule
%       &  RMSE & Bias & WMAPE & Ranking \\ \midrule
%     A-CNN & 19.67 & -0.31 & 16.47 & 3  \\ 
%     CNN & 22.96 & 0.06 & 18.49 & 6 \\
%     A-ConvLSTM & 18.95 & 0.12 & 15.51 & 1 \\
%     ConvLSTM & 19.05 & -0.20 & 15.10 & 2 \\
%     A-GCN-LSTM & 22.09 & 0.52 & 17.25 & 5 \\ 
%     GCN-LSTM & 21.80 & 0.03 & 17.51 & 4 \\  \bottomrule
%     \end{tabular}
%   }
% \end{table}

\begin{figure*}[tbp]
  \centering
  \includegraphics[width=1\textwidth]{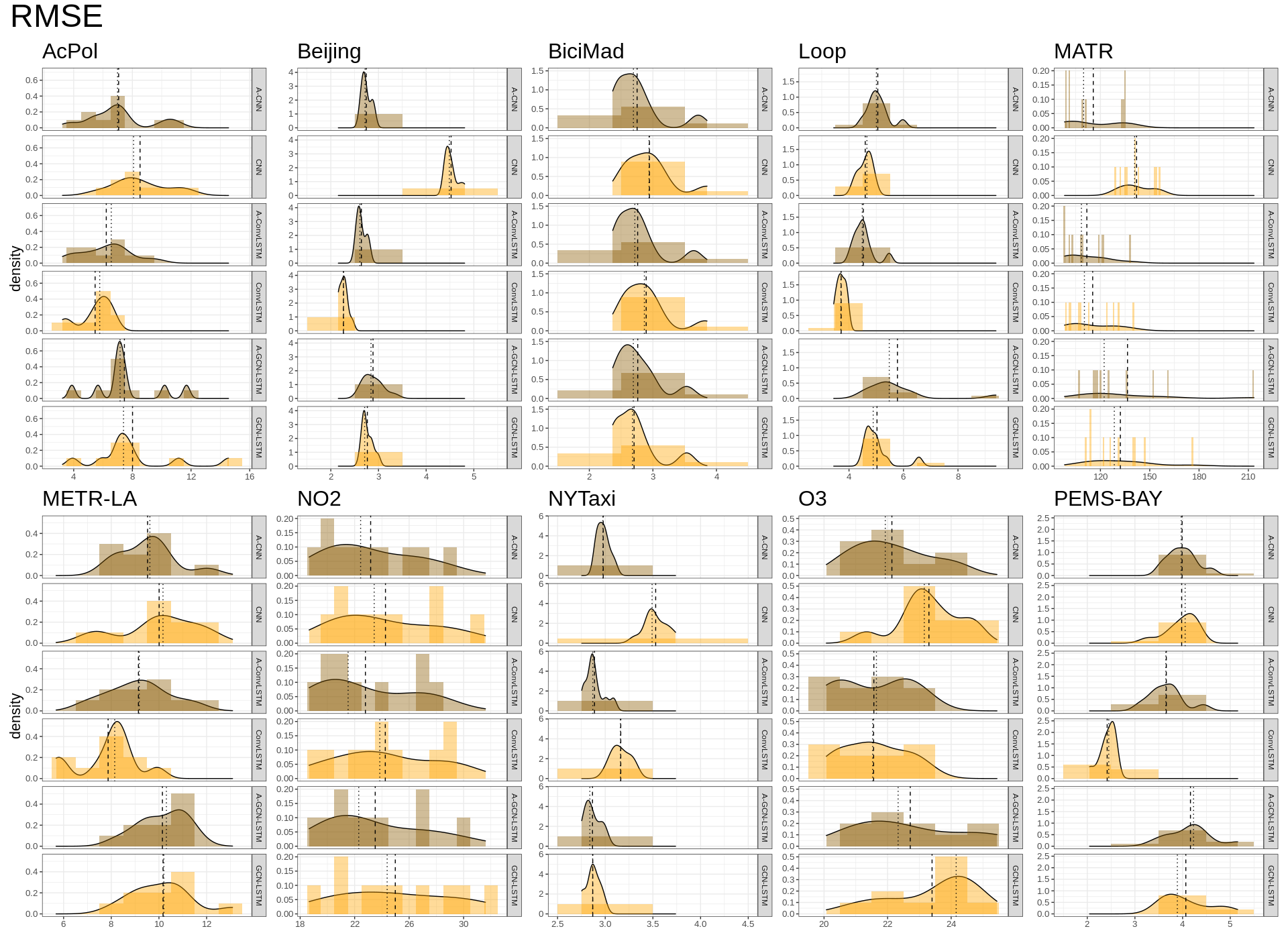}
  \caption{RMSE distribution for each model and dataset. Dashed vertical line
    represents the mean, dotted vertical line represents median.}
  \label{fig:res_1}
\end{figure*}

First of all, we can verify the goodness of our experiments by direct comparison
with analogous studies \cite{de_medrano_spatio-temporal_2020,
  navares_predicting_2020, cui_traffic_2019, bbliaojqZhangKDD18deep,
  li_diffusion_2018}, showing that our results are in line with them. Since most
of the datasets have already been used, we can extrapolate this idea to those
which have not. To better visualize the error over all datasets, Fig.
\ref{fig:res_1} shows RMSE distribution. From this figure we can deduct that, in
general terms, agnostic models show a similar behavior than their respective
main competitors.

The performance of the different strategies over individual datasets is directly
associated with the spatial autocorrelation metric in Table
\ref{tab:statistics}. On the one hand, datasets with a higher value of Moran's
$I$ have a propensity to show better performance with traditional models
(Beijing, Loop, METR-LA, and PEMS-BAY). On the other hand, datasets with lower
values of the same metric usually show better behavior with the agnostics
versions (AcPol, BiciMad, MATR, NO2, NYTaxi, and O3).
 
% In table \ref{tab:res_average} we can see average performance for each model
% over all datasets, and the resulting ranking. 
In order to inquire into these results and
provide statistical evidence, a Friedman rank test was performed over the error
distribution for all datasets. A Friedman statistic of $F = 21.6$, distributed
according to a $\chi^{2}$ with 5 degrees of freedom obtains a p-value of
$6.2e-4$ with $\alpha = 0.05$, which provides evidence of the existence of a
significant difference between the algorithm.

Given that Friedman's null hypothesis was rejected, a post-hoc pairwise
non-parametric based comparison was carried out to check the differences between
the proposed algorithms with Holm and Benjamini-Hochberg adjustments. As we are
especially interested in testing whether the introduction of spatial information
as a \textit{prior} is necessary or not, Table \ref{tab:res_p} shows statistical
significance in the traditional-agnostic model comparison for all datasets.
Through these tests we compare if there are significant differences between
the means of two different algorithms error distributions. Thus, for each hypothesis
the test accepts or rejects the idea that the two models that compose the hypothesis
generate, statistically speaking, same error distributions. By looking at 
this table we can confirm our initial claim since there is not enough evidence to
support that traditional methods suppose an
improvement over their agnostic versions. In fact, the only comparison that yields
a significant result (hypothesis I) show evidence in favor of the agnostic model.

\begin{table}[btp]
\centering
\caption{Adjusted Holm and Benjamini-Hochberg $p$-values with pairwise rejected
  hypothesis at $\alpha = 0.05$ for all datasets. A $p$-value lower than $\alpha$
  suggest that both algorithms produce different error
  distributions. % A p-value bigger than $\alpha$ reject the hypothesis that both
  % algorithms produce different error distributions.
}
  \label{tab:res_p}
  \scalebox{0.85}{
\begin{tabular}{cllll}
  \hline
 i & hypotheses & $p_{\mathrm{unajusted}}$ & $p_{\mathrm{holm}}$ & $p_{\mathrm{BH}}$ \\ 
  \hline
  I & A-CNN vs CNN & 0.014 & 0.048 & 0.023   \\ 
  II &  A-ConvLSTM vs ConvLSTM & 0.77  & 1  & 0.825  \\ 
  III &  A-GCN-LSTM vs GCN-LSTM & 0.736  & 1  & 0.846  \\ 
   \hline
\end{tabular}
}
\end{table}

In terms of computational performance, Table \ref{tab:res_time} summarises
average run times per fold, model, and dataset, and the number of parameters per
dataset for all models (recall that, to facilitate a fairer comparison, all
models have the same number of parameters for every problem, see Section
\ref{S4.3}). Again, no differences are reported between traditional and agnostic
models neither. As we would expect, A-CNN and CNN models show great advantage in
terms of time consumption compared to the rest of the methodologies.
% \jnote{Es contraintuitivo que todos los modelos tengan el mismo número de parámetros.
%   Aunque la explicación está más arriba, convendría explicarlo aquí brevemente.}

\begin{table*}[btp]
  \centering
  \caption{Average run time per fold in secods and approximate number of parameters used per dataset.}
  \label{tab:res_time}
  \scalebox{0.7}{
  \begin{tabular}{lccccccccccl}
    \toprule
        &  AcPol & Beijing & BiciMad & LOOP & MATR & METR-LA & NO2 & NYTaxi & O3 & PEMS-BAY & Average\\ \midrule
    A-CNN & 1.0 & 16.7 & 4.1 & 97.7 & 26.7 & 36.5 & 56.8 & 3.6 & 42.8 & 67.9 & 36.3 \\ 
    CNN & 1.4 & 8.1 & 13.4 & 117.5 & 44.5 & 68.6 & 33.2 & 18.4 & 22.5 & 48.0 & 37.8 \\
    A-ConvLSTM & 4.8 & 41.9 & 5.1 & 229.1 & 411.8 & 179.6 & 67.8 & 9.2 & 56.2 & 349.0 & 135.5 \\
    ConvLSTM & 2.6 & 103.2 & 38.2 & 350.0 & 422.5 & 238.7 & 74.8 & 13.0 & 56.8 & 400.7 & 171.6 \\
    A-GCN-LSTM & 18.1 & 72.6 & 15.0 & 168.7 & 146.0 & 83.7 & 449.4 & 24.4 & 221.2 & 172.3 & 137.1 \\ 
    GCN-LSTM & 13.3 & 71.9 & 12.5 & 190.5 & 111.7 & 77.9 & 467.4 & 30.5 & 221.6 & 199.9 & 141.0 \\ \midrule
    Number of parameters & $\sim$ 50K & $\sim$ 200K & $\sim$ 150K & $\sim$ 250K & $\sim$ 200K & $\sim$ 150K & $\sim$ 200K & $\sim$ 150K & $\sim$ 150K & $\sim$ 250K \\ \bottomrule
  \end{tabular}
  }
  
\end{table*}

\subsection{Spatial agnosticism: permutation test}
\label{S5.2}

To further validate one the most important statements of this work, i.e. to ensure 
that the models we have presented as spatially agnostic really are, we propose to
randomly permutate the spatial dimension of data before training. As we just
want to compare the behavior of the different methods when input data is not
sorted, we are only interested in studying how the error distributions are
modified when this perturbation is introduced in the system, and not in pure
performance. Given that ConvLSTM and A-ConvLSTM have shown to be a statistically
significant better option than the other models, we will
use only these two models through this section. 

\begin{figure*}[htbp]
\centering
\includegraphics[width=1\textwidth]{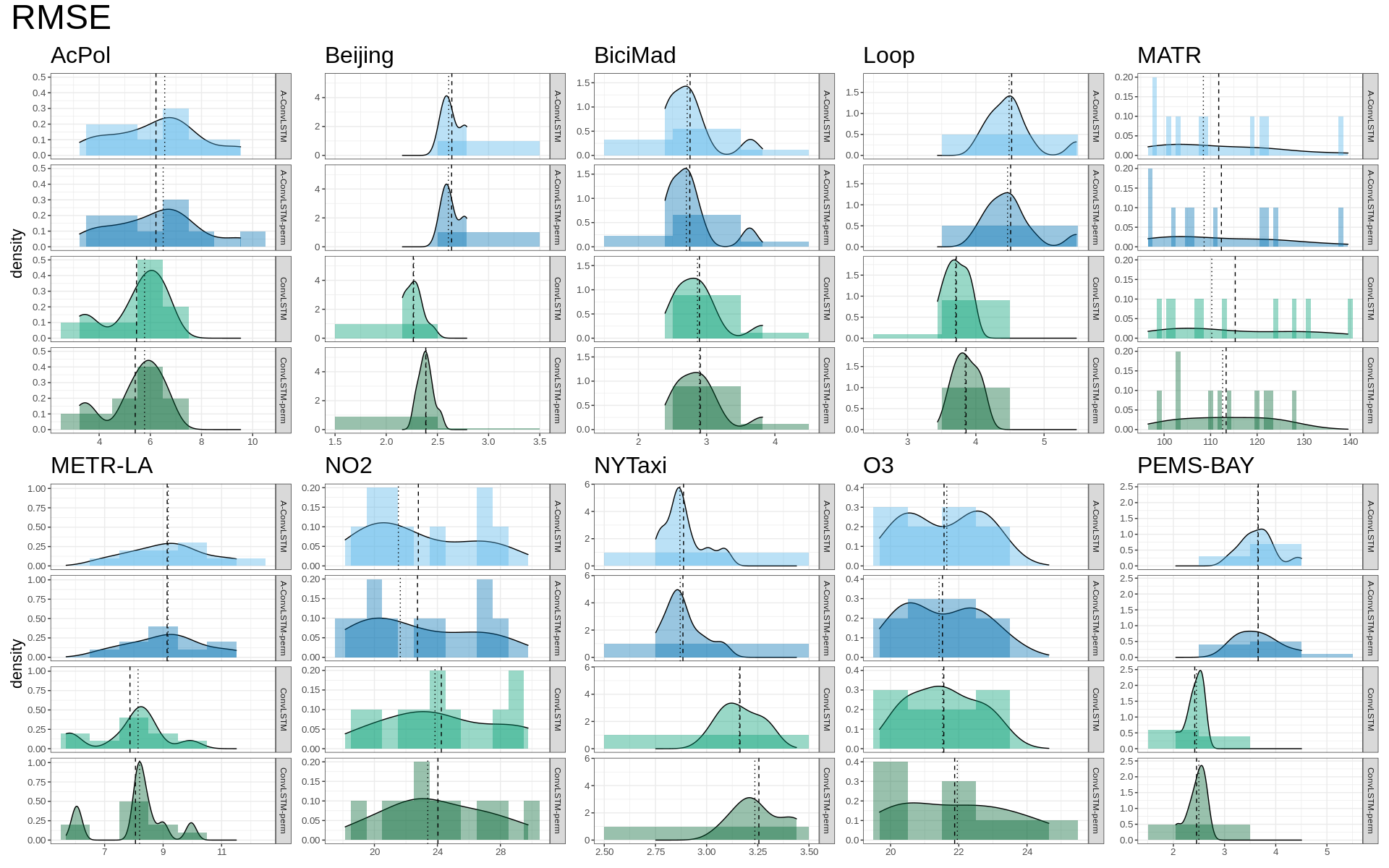}
\caption{RMSE distribution for each model and dataset before and after training
  with randomly permutations in their spatial dimension. Dashed vertical line
  represents the mean, dotted vertical line represents median. In blue,
  A-ConvLSTM-based models and in green ConvLSTM-based models.}
\label{fig:res_2}
\end{figure*}

In Fig. \ref{fig:res_2} we can visualize the RMSE results for both models before
and after (\textit{model name-perm}) the random permutation.

From this last figure we can clearly see that error distributions for A-ConvLSTM
and A-ConvLSTM-perm are practically identical for all the problems, while that
does not happen for ConvLSTM and ConvLSTM-perm. Thus, we carry out a post-hoc
pairwise non-parametric based comparison to check the differences between the
models with Holm and Benjamini-Hochberg adjustments. Table \ref{tab:res_perm_p}
shows the aforementioned $p$-values, marking with asterisks (*) those that are
statistically significant. In the table, ``A-ConvLSTM'' refers to the comparison
A-ConvLSTM vs A-ConvLSTM-perm, while ``ConvLSTM'' refers to the comparison
ConvLSTM vs ConvLSTM-perm.

\begin{table*}[btp]
\caption{Adjusted Holm and Benjamini-Hochberg $p$-values with pairwise rejected hypothesis at 
$\alpha = 0.05$ for all datasets after testing spatial agnosticism via random
permutation. Rejected hypothesis (meaning both algorithms produce different
error distributions) are marked with *.
% A p-value lower than $\alpha$ (*) accepts the hypothesis that both algorithms produce
% different error distributions. A p-value bigger than $\alpha$
% reject the hypothesis that both algorithms produce different error
% distributions.
}
  \label{tab:res_perm_p}
\resizebox{\textwidth}{!} {
\begin{tabular}{@{}llllllllllllllll@{}}
\toprule
    & \multicolumn{3}{l}{AcPol} & \multicolumn{3}{l}{Beijing} & \multicolumn{3}{l}{BiciMad} & \multicolumn{3}{l}{LOOP} & \multicolumn{3}{l}{MATR} \\ \cmidrule(l){2-4}
\cmidrule(l){5-7}
\cmidrule(l){8-10}
\cmidrule(l){11-13}
\cmidrule(l){14-16}
           & $p_{\mathrm{unajusted}}$ & $p_{\mathrm{holm}}$ & $p_{\mathrm{BH}}$     & $p_{\mathrm{unajusted}}$ & $p_{\mathrm{holm}}$ & $p_{\mathrm{BH}}$     & $p_{\mathrm{unajusted}}$ & $p_{\mathrm{holm}}$ & $p_{\mathrm{BH}}$     & $p_{\mathrm{unajusted}}$ & $p_{\mathrm{holm}}$ & $p_{\mathrm{BH}}$    & $p_{\mathrm{unajusted}}$ & $p_{\mathrm{holm}}$ & $p_{\mathrm{BH}}$    \\ \midrule
A-ConvLSTM    &  0.922 & 1  & 0.922  & 0.193     & 0.193    & 0.193    & 0.496     & 0.496   & 0.496      & 0.027*  & 0.027*     & 0.027*    & 0.232  & 1  & 0.668 \\
ConvLSTM   &  0.922 & 1  & 0.922  & 0.014*     & 0.027*    & 0.016*    & 0.027*     & 0.027*   & 0.027*      & 0.002*  & 0.012*     & 0.002*    & 0.557  & 1  & 0.668 \\ \bottomrule
\end{tabular}
}
%\end{table}
  \\\vspace{0.1cm}\\
%\begin{table}[hbtp]
%  \caption{\jnote{Falta la descripción de esta tabla. Además, creo que es
%      preferible ubicar los ``floats'' con 'tbp' o 'btp', en vez de 'hbtp'.}}
  \resizebox{\textwidth}{!} {
    \begin{tabular}{@{}llllllllllllllll@{}}
      \toprule
          & \multicolumn{3}{l}{METR-LA} & \multicolumn{3}{l}{NO2} & \multicolumn{3}{l}{NYTaxi} & \multicolumn{3}{l}{O3} & \multicolumn{3}{l}{PEMS-BAY} \\ \cmidrule(l){2-4}
      \cmidrule(l){5-7}
      \cmidrule(l){8-10}
      \cmidrule(l){11-13}
      \cmidrule(l){14-16}
           & $p_{\mathrm{unajusted}}$ & $p_{\mathrm{holm}}$ & $p_{\mathrm{BH}}$     & $p_{\mathrm{unajusted}}$ & $p_{\mathrm{holm}}$ & $p_{\mathrm{BH}}$     & $p_{\mathrm{unajusted}}$ & $p_{\mathrm{holm}}$ & $p_{\mathrm{BH}}$     & $p_{\mathrm{unajusted}}$ & $p_{\mathrm{holm}}$ & $p_{\mathrm{BH}}$    & $p_{\mathrm{unajusted}}$ & $p_{\mathrm{holm}}$ & $p_{\mathrm{BH}}$    \\ \midrule
      A-ConvLSTM      & 0.777   & 0.777   & 0.777   & 0.777      & 0.984      & 0.777      & 0.375     & 0.375     & 0.375     & 0.557    & 1     & 1      & 0.846    & 0.846 & 0.846  \\
      ConvLSTM        & 0.011*   & 0.02*    & 0.012*   & 0.492  & 0.984  & 0.59   & 0.01*      & 0.022*     & 0.013*     & 0.846    & 1         & 0.846     & 0.004*    & 0.012* & 0.005*  \\ \bottomrule
    \end{tabular}
  }
\end{table*}

This table lets us conclude that A-ConvLSTM shows spatial agnosticism and its
performance is unaffected by how the spatial dimension is treated. However, the
ConvLSTM presents an important discrepancy in terms of performance when
unsorting the grid. Although this premise holds in general terms over all
datasets, it can be seen again that the results are directly related to
correlation metrics in Table \ref{tab:statistics}: those datasets with a higher
value of Moran's $I$ tend to suffer more with the permutation test (Beijing,
LOOP, METR-LA, and PEMS-BAY). As in those cases the spatial autocorrelation is
higher, sharing parameters in the spatial dimension is more beneficial, and
changing the grid has a greater effect.

\subsection{Practical guidelines}
\label{S5.3}

Given our results, we can provide some guidelines in order to help other
practitioners working with real spatio-temporal problems:
% \jnote{Si son guidelines, tienen que ser más propositivas.}
\begin{itemize}
\item %Experimental evidence over a set of real datasets strongly suggest that
  Assuming neighborhood-based relations as a premise when approaching a
  spatio-temporal problem with neural networks might not always be the best
  option. Instead of naively assuming these spatial relations, it might be
  beneficial to dig more deeply in the data analysis or to rethink how the
  problem is addressed.
  %The results of these experiments and its integration with autocorrelation
  %metrics confirm our main hypothesis:
  Concretely, real datasets do not necessarily are similar through spatial locations,
  contrary to what is usually assumed. Thus, the nature of data should be
  reflected when defining the network architecture. In any case, further
  considerations should be given to preliminaries studies of the spatial
  distribution of the data.
%   \jnote{No entiendo qué se propone exactamente en este
%     punto. ¿Estudiar los datos? ¿No es eso lo que se propone en el anterior? Si
%     es así, fusionaría ambos.}
\item When the distribution of the data shows a clear spatial relationship based
  on neighborhood, as in the case of large traffic sensor networks, the
  traditional format of convolution-based networks might be advantageous.
  However, when this is not clearly verified, as for example with air quality,
  models do not show improvement by sharing weights between different locations.
\item If there is not enough available evidence about the spatial distribution
  characteristics, spatially agnostic models might be best suited as they are
  capable of performing well while being less laborious to work with.
\item In any case, consider using spatial agnostic models if your needs in terms
  of precision must be balanced with the available resources.% throughout our experiments they have shown to be able to behave well in a wide variety of problems.
\end{itemize}

% %%%%%%%%%%% %
% CONCLUSIONS %
% %%%%%%%%%%% %
\section{Conclusions}
\label{S6}

Through this work, we have explored how classical spatial assumptions based on
closeness are not always the best deal when working with convolutional neural
networks for spatio-temporal series regression. Due to their usual lack of spatial
autocorrelation, other alternatives might be more suited. In order to test this
idea, we have compared several versions of convolutional-based models that make
no use of \textit{prior} spatial information (neither directly nor indirectly),
namely spatial agnostic, with their respective traditional forms. Spatial
agnostic models are a perfect tool to contrast our hypothesis as they do not use
extra modules or steps as others, but tackle the problem directly purely via
convolutions.

After extensive and standardized experimentation, we can confirm our main
hypothesis: the inclusion of adjacency-based representations of the spatial
distribution of real data does not necessarily fit well for the classical
convolutional shared-weights scheme. Concretely, without using any specific
spatial mechanism, spatial agnostic models have been shown to be equal in
performance to some of the most notable spatio-temporal models. Also, we have
shown how these models, unlike traditional convolutional methods, are really
spatially agnostic, and how this is related to the spatial autocorrelation of
the series. Furthermore, beyond proving our initial hypothesis we have shown how
our methodology is simpler and less laborious to work with, offering the
possibility of obtaining good performance without having to carry out extra
research about the application domain. Finally, by analyzing ten different
datasets with different spatio-temporal conditions each, we can confirm the
statistical significance of these statements with a confidence of $95\%$.

\section{Acknowledgements}
\label{S7}

This research has been partially funded by the \textit{Empresa Municipal de
  Transportes} (EMT) of Madrid under the chair \textit{Aula Universitaria
  EMT/UNED de Calidad del Aire y Movilidad Sostenible}.

% \begin{IEEEbiographynophoto}{Rodrigo de Medrano}
% Rodrigo de Medrano received his B.S. (2018) in physics from the UAM university in Madrid (Spain), and the M.S. (2019) in Artificial Intelligence at the UNED university. He is currently working on his Ph.D. at the Department of Artificial Intelligence of the UNED university. His research interests include traffic safety and forecasting, air quality and sustainable transportation.
% \end{IEEEbiographynophoto}

% \begin{IEEEbiographynophoto}{José Luis Aznarte}
%   José Luis Aznarte is an associate professor at the Department of Artificial
%   Intelligence of the UNED university. He was a post-doc researcher with the
%   Renewable Energy Research Group in the Center for Energy and Processes of the
%   MINES ParisTech engineering school (France), where he was engaged in the
%   European projects ANEMOS.Plus (FP6) and Safewind (FP7). In 2013, he was
%   awarded a Ramón y Cajal tenure track. Nowadays he coordinates research about
%   applications of data mining, machine learning and soft computing, especially
%   in the framework of time series forecasting. He is involved in the development
%   of operational solutions for air quality forecasting, amongst other
%   applications.
% \end{IEEEbiographynophoto}

\bibliographystyle{./bibliography/IEEEtran}
\bibliography{./bibliography/bibliography.bib}

\end{document}